\documentclass[journal]{IEEEtai}

\usepackage[colorlinks,urlcolor=blue,linkcolor=blue,citecolor=blue]{hyperref}

\usepackage{color,array}
\usepackage{graphicx}
\usepackage{subcaption}
\usepackage{placeins}
\usepackage{siunitx}
\usepackage{orcidlink}

\usepackage{amsmath,amssymb,amsfonts} 
\usepackage{mathtools}                
\usepackage{bm}                       

\usepackage{physics}                  

\usepackage[ruled,vlined]{algorithm2e}
 \usepackage{tikz}
\usetikzlibrary{arrows.meta,positioning}

\begin{document}

\title{Lane-Frame Quantum Multimodal Driving Forecasts for the Trajectory of Autonomous Vehicles}

\author{
    Navneet Singh \orcidlink{0000-0002-3940-1608}~\textit{{Student Member,~IEEE}} and 
    Shiva Raj Pokhrel \orcidlink{0000-0001-5819-765X}~\textit{{Senior~Member,~IEEE}} 
    \thanks{
    Authors are with the IoT \& Software Engineering Research Lab, School of Information Technology, Deakin University, Geelong, VIC 3125, Australia (e-mail: \href{mailto:n.navneetsingh@deakin.edu.au}{n.navneetsingh@deakin.edu.au}; \href{mailto:shiva.pokhrel@deakin.edu.au}{shiva.pokhrel@deakin.edu.au}).\\
    }
}



\maketitle
\begin{abstract}
Trajectory forecasting for autonomous driving must deliver accurate, calibrated multi-modal futures under tight compute and latency constraints. We propose a compact hybrid quantum architecture that aligns quantum inductive bias with road-scene structure by operating in an ego-centric, lane-aligned frame and predicting residual corrections to a kinematic baseline instead of absolute poses. The model combines a transformer-inspired quantum attention encoder (9 qubits), a parameter-lean quantum feedforward stack (64 layers, ${\sim}1200$  trainable angles), and a Fourier-based decoder that uses shallow entanglement and phase superposition to generate 16 trajectory hypotheses in a single pass, with mode confidences derived from the latent spectrum. All circuit parameters are trained with Simultaneous Perturbation Stochastic Approximation (SPSA), avoiding backpropagation through non-analytic components. In the Waymo Open Motion Dataset, the model achieves minADE (minimum Average Displacement Error) of \SI{1.94}{m} and minFDE (minimum Final Displacement Error) of \SI{3.56}{m} in the $16$ models predicted over the horizon of \SI{2.0}{s}, consistently outperforming a kinematic baseline with reduced miss rates and strong recall. Ablations confirm that residual learning in the lane frame, truncated Fourier decoding, shallow entanglement, and spectrum-based ranking focus capacity where it matters, yielding stable optimization and reliable multi-modal forecasts from small, shallow quantum circuits on a modern autonomous-driving benchmark.
\end{abstract}



\section{Introduction}

\IEEEPARstart{A}{ccurate} short-horizon trajectory forecasting under uncertainty is a central requirement for autonomous driving. An effective forecaster must reason about multiple plausible futures (e.g., straight vs.\ turn), remain well-calibrated, and operate under strict latency and compute budgets. Classical deep models achieve high accuracy~\cite{teng2023motion,gu2021densetnt} but often at the cost of large parameter counts, multi-pass decoding, and heavy training pipelines~\cite{ettinger2021large,casas2021mp3}. This paper explores a different point in the design space: \emph{can a compact, shallow hybrid quantum model, carefully aligned with the structure of near-term driving, produce useful multi-modal forecasts under tight resource constraints?}

Rather than aiming for state-of-the-art performance or formal quantum advantage, we present an \emph{early, small-scale feasibility study} that examines the applicability of quantum machine learning to vehicle trajectory prediction. Our goal is to understand whether a modest, task-structured quantum circuit, with only a handful of qubits and shallow depth, can participate meaningfully in a modern forecasting pipeline when equipped with strong classical inductive biases.
\begin{figure*}[t]
  \centering
  \includegraphics[width=\textwidth]{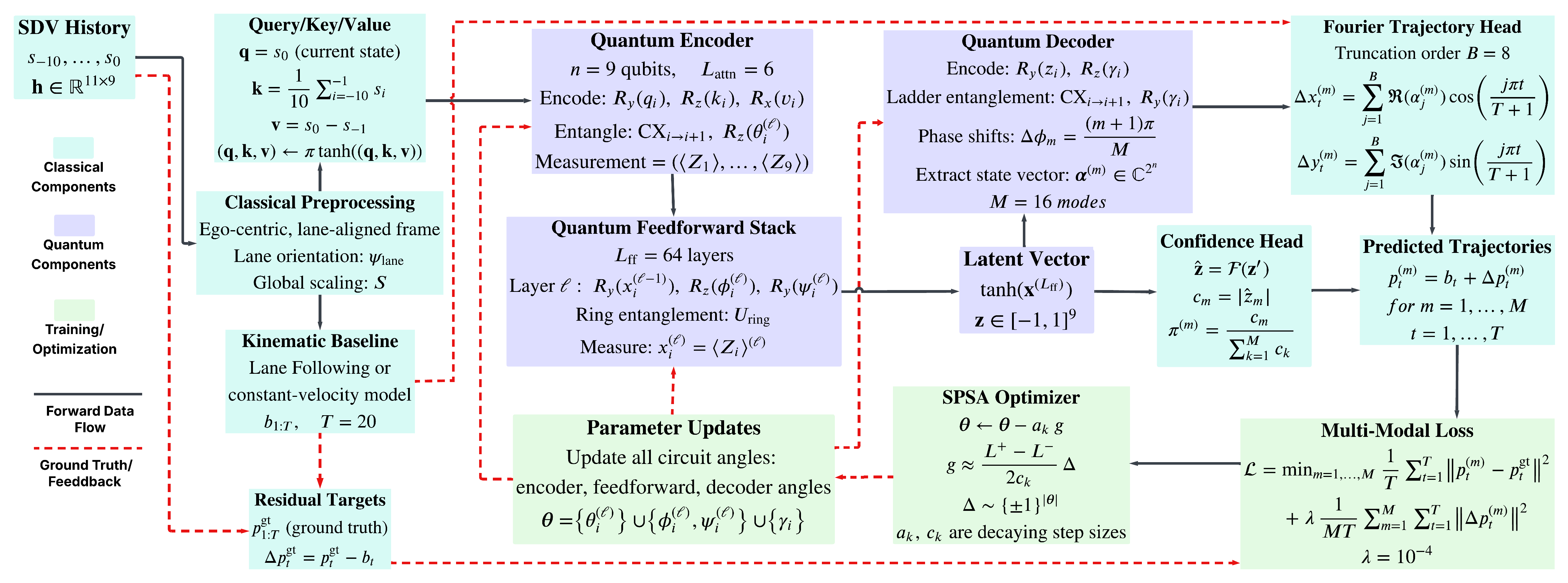}
  \caption{Overview of the proposed quantum multi-modal trajectory prediction pipeline. SDV history is preprocessed into an ego-centric lane frame and split into a kinematic baseline and query-key-value features, which are encoded by a quantum attention encoder and a deep quantum feedforward stack to produce a latent vector. A quantum decoder, Fourier trajectory head, and confidence head generate multi-modal residual trajectories and mode probabilities that are added to the baseline to obtain final predictions. Solid arrows denote forward data flow, while dashed arrows show construction of residual targets from ground truth and the SPSA-based parameter updates used for training.}
  \label{fig:flow_fig}
\end{figure*}

We introduce a small, task-aligned hybrid quantum architecture that deliberately reduces what the model needs to represent as shown in Fig~\ref{fig:flow_fig}. Each scenario is first expressed in an ego-centric, lane-aligned coordinate frame and paired with a simple kinematic/map-following baseline~\cite{paden2016survey}. The model then predicts only \emph{residual} motion relative to this baseline, shifting learning from full trajectories to smooth, meter-scale corrections that match the statistics of short-horizon driving~\cite{wang2022ltp}. This choice contracts the output dynamic range, stabilizes optimization, and focuses capacity where it matters~\cite{long2025physical}.

The model comprises three lightweight quantum components. A \emph{quantum attention encoder} (on 9 qubits) maps query-key-value summaries of the recent motion into an entangled state, providing transformer-inspired mixing of current pose, historical trend, and short-term change using linear/ladder entanglement~\cite{chen2025quantum}. A \emph{deep yet parameter-lean quantum feedforward stack} (64 layers, \(\sim\!1200\) trainable angles overall) alternates data re-injection, local rotations, and ring entanglement, with measurements after each layer to form a stable hybrid loop~\cite{broughton2020tensorflow}. Finally, a \emph{Fourier-based quantum decoder} uses phase superposition to emit \(M{=}16\) coherent trajectory hypotheses in a \emph{single} forward pass; a fixed confidence mapping derived from the latent spectrum ranks these hypotheses without an additional learned head.

Design choices embed strong inductive biases. Residual learning in the lane frame confines the circuit to small, structured deviations from a map/CTRV prior. The truncated Fourier readout imposes a low-frequency smoothness bias appropriate for 2.0\,s horizons, while global phase shifts generate diverse, geometrically consistent futures at constant inference cost in \(K\). Shallow entanglement keeps the parameter space small and the Simultaneous Perturbation Stochastic Approximation (SPSA) landscape well-conditioned. Together, these constraints allow a 9-qubit circuit to behave like a larger model \emph{in-distribution}, trading universal expressivity for targeted efficacy.

We train the system end-to-end with SPSA, which is well-suited to parameterized quantum circuits and to our non-differentiable '\(\min\) over modes' objective. On the Waymo Open Motion Dataset (WOMD)~\cite{ettinger2021large}, using standard evaluation on a held-out subset, the model attains meter-scale displacement errors, broad hypothesis coverage, and stable calibration, despite the small qubit count, shallow depth, and fixed-basis decoder. Miss rates and recall/precision-recall trends improve consistently over training, and the optimization trajectory exhibits smooth loss reduction, shrinking variance, and a staircase of best-so-far improvements, indicating predictable convergence under SPSA. Compared to a strong kinematic baseline, the model substantially reduces displacement errors while maintaining single-pass multi-modality.

This work sits adjacent to, but distinct from, classical Transformers. Our encoder is \textit{transformer-inspired} (attention-like mixing realized by quantum gates), yet the overall system is not a general-purpose \textit{quantum Transformer.} Instead, it is a narrowly scoped, task-structured hybrid architecture that exploits superposition and interference to generate and rank multiple futures efficiently. 


\paragraph{Key Contributions}
\begin{itemize}
\item \textit{A compact, task-aligned hybrid quantum forecaster.} We combine a quantum attention encoder, a parameter-lean feedforward stack, and a Fourier decoder that leverages phase superposition to produce \(M{=}16\) modes in one pass.
\item \textit{Residual learning in a lane-aligned frame.} Predicting corrections atop a map/CTRV baseline reduces dynamic range and centers learning on lane-consistent adjustments, improving stability and data efficiency.
\item \textit{Single-pass multi-modality with physics-shaped ranking.} Phase-offset decoding yields diverse, smooth hypotheses at constant cost, while a latent-spectrum confidence assigns meaningful ranks without an extra learned head.
\item \textit{Reliable gradient-free training for multi-modal objectives.} We demonstrate smooth convergence and shrinking variance when optimizing shallow circuits with SPSA under a non-smooth 'min-over-modes' loss.
\item \textit{An early feasibility study of quantum trajectory forecasting on WOMD.} Using only 9 qubits and shallow entanglement, the model achieves meter-scale accuracy on a representative WOMD subset, outperforming a kinematic baseline and illustrating that even very small quantum circuits can contribute to multi-modal vehicle trajectory prediction.
\end{itemize}

\subsection{Literature}
The literature on trajectory prediction is extensive, so we focus on work most closely aligned with our setting and design choices.

Early deep-learning approaches treat motion prediction as sequence modeling over agent states, using LSTMs for pedestrians and vehicles. Social-LSTM~\cite{alahi2016social} explicitly encodes nearby agents via social pooling to forecast human trajectories in crowds, while Lin et al.\ use LSTMs with attention for human-driven vehicle longitudinal motion in connected environments~\cite{lin2021long}. Surveys on human and vehicle trajectory forecasting emphasize that realistic predictions must jointly account for agent interactions, map structure and uncertainty, especially for downstream motion planning~\cite{kothari2021human,huang2023review,liu2021survey}. Graph-based models such as Trajectron~\cite{ivanovic2019trajectron} and Trajectron++~\cite{salzmann2020trajectron} represent multi-agent scenes as dynamic spatiotemporal graphs and output probabilistic, multi-modal trajectories that respect vehicle dynamics and can incorporate heterogeneous inputs. Anchor- and goal-based approaches such as TNT~\cite{zhao2021tnt} and MultiPath++~\cite{varadarajan2022multipath} further improve diversity and accuracy by predicting a compact set of endpoints or latent anchors together with associated trajectories on large-scale driving datasets, including Argoverse and WOMD. More recently, transformer architectures Scene Transformer~\cite{ngiam2021scene}, HiVT~\cite{zhou2022hivt}, Wayformer~\cite{nayakanti2022wayformer} and MotionLM~\cite{seff2023motionlm} use global attention over agents, time and HD-map polylines to achieve state-of-the-art performance on WOMD and related benchmarks, but they rely on very deep networks with millions of parameters and substantial inference latency, which complicates real-time deployment and large-scale calibration.

In parallel, hybrid data–model driven frameworks encode vehicle physics and driver heterogeneity within deep architectures. Bai et al.\ propose KO-TAFTN, which couples Kepler-based physical modeling of driver-specific preferences with temporal attention over agents and planned CAV trajectories to capture anticipatory interactions in mixed highway traffic~\cite{bai2025hybrid}. Using variable selection and interpretable attention, KO-TAFTN produces diverse maneuver-level predictions and outperforms purely data-driven baselines, underscoring the value of structured priors in high-capacity sequence models.

Quantum neural networks (QNNs)~\cite{singh2025modeling,11231118} based on parameterized quantum circuits (PQCs) have emerged as an alternative paradigm for sequence modeling, aiming to obtain high expressivity with relatively few trainable parameters. Chen et al.\ introduce Quantum Long Short-Term Memory (QLSTM), a hybrid model that embeds variational quantum circuits inside LSTM cells and shows competitive or faster convergence than classical LSTMs on temporal benchmarks with shallow circuits suitable for NISQ devices~\cite{chen2022quantum}. Takaki et al.\ propose a variational quantum recurrent neural network (VQRNN) that learns temporal dependencies via PQCs arranged in a recurrent structure and accurately models classical and quantum dynamical time series with a compact number of qubits and parameters~\cite{takaki2021learning}.

Existing motion-forecasting systems for autonomous driving therefore fall broadly into: (i) large classical architectures (GNNs, transformers, and hybrid physics–data models such as KO-TAFTN) that achieve high accuracy on WOMD and highway benchmarks but incur significant computational and memory cost, and (ii) emerging quantum and hybrid quantum–classical models that show promise on generic or small-scale time-series tasks yet have not been evaluated on realistic, map-conditioned SDV trajectory prediction benchmarks. To our knowledge, there remains a gap in systematically assessing shallow, few-qubit PQC-based models on short-horizon SDV trajectory forecasting in WOMD, with an emphasis on calibrated multi-hypothesis outputs under tight compute budgets. Our work addresses this gap by formulating lane-aligned residual prediction as a compact quantum sequence modeling problem, designing a low-depth hybrid QNN over 9-dimensional SDV state histories, and empirically comparing its accuracy and calibration against classical baselines in a deployment-oriented setting.

\section{Methodology}
\subsection{Data Preprocessing}
\begin{algorithm}[t]
\caption{Construction of training examples and residual targets}
\label{alg:data}
\DontPrintSemicolon
\KwIn{Waymo scenarios; map; horizon $T{=}20$; scale $S$}
\KwOut{triplets $(\mathbf{h}, b_{1:T}, \Delta p^{\mathrm{gt}}_{1:T})$}

examples $\leftarrow$ empty list\;

\ForEach{scenario and valid SDV index $t_0$}{
  $(s_{-10},\dots,s_0, p^{\mathrm{gt}}_{1:T}) \leftarrow \textsc{ExtractStates}(\text{scenario},t_0)$\;
  $\psi_{\text{lane}} \leftarrow \textsc{LaneOrientation}(s_0,\text{map})$\;
  $(s_{-10},\dots,s_0, p^{\mathrm{gt}}_{1:T}) \leftarrow \textsc{ToLaneFrame}(s_{-10},\dots,s_0, p^{\mathrm{gt}}_{1:T}, \psi_{\text{lane}}, S)$\;

  $v \leftarrow \textsc{Speed}(s_0)$\;
  $\omega \leftarrow \textsc{YawRate}(s_0)$\;

  \eIf{$\|v\|\ge 0.05$ and $\textsc{LaneAvailable}(\text{map})$}{
    $b_{1:T} \leftarrow \textsc{LaneFollowingBaseline}(s_0,\text{map},v,T)$\;
  }{
    $b_{1:T} \leftarrow \textsc{KinematicBaseline}(s_0,v,\omega,T)$\;
  }

  \For{$t \leftarrow 1$ \KwTo $T$}{
    $\Delta p_t^{\mathrm{gt}} \leftarrow p^{\mathrm{gt}}_t - b_t$\;
  }

  $\mathbf{h} \leftarrow \textsc{StackStates}(s_{-10},\dots,s_0)$\;
  append $(\mathbf{h}, b_{1:T}, \Delta p^{\mathrm{gt}}_{1:T})$ to examples\;
}
\Return{all stored $(\mathbf{h}, b_{1:T}, \Delta p^{\mathrm{gt}}_{1:T})$}\;
\end{algorithm}

We use the publicly available WOMD, which provides $\sim$100k driving scenarios (20\,s at 10\,Hz) with tracked trajectories and high-definition maps. Each scenario contains a designated self-driving vehicle (SDV, the ego) and other traffic agents. We focus on predicting the SDV’s future trajectory from its past motion and map context. The complete preprocessing pipeline, including lane-aligned coordinates, the kinematic baseline $b_{1:T}$ and residual targets $\Delta p^{\mathrm{gt}}_{1:T}$, is summarized in Algorithm~\ref{alg:data}.

The SDV state is represented by 11 time steps (10 past at 0.1\,s plus the current step at time $t_0$). Each time step has a 9-dimensional feature vector: position $(x,y,z)$, ground-plane velocity $(v_x,v_y)$, yaw rate $\omega$, heading angle, and vehicle dimensions (length and width). We denote the history as $\{s_{-10},\dots,s_{-1},s_0\}$, where $s_0$ is the current state. The prediction target is the SDV trajectory for the next $T=20$ steps. 

We first transform each scenario into an ego-centric, lane-aligned frame. The SDV’s current position becomes the origin, and a reference yaw $\psi_{\text{lane}}$ is computed from the map by averaging local road direction vectors within a radius around the SDV. Positions are rotated so that the $x$-axis aligns with this lane direction and translated so that $(0,0)$ corresponds to the SDV at $t_0$. Velocities are transformed similarly.

Next, we construct a simple kinematic baseline trajectory for the SDV. If the SDV speed is non-negligible ($\ge 0.05$\,m/s), we simulate motion for $T$ steps along the map’s local road direction at the current speed. If the vehicle is turning (non-zero yaw rate), the baseline follows the road curvature; if reliable map vectors are unavailable, we instead use the current heading. For very low speed or ambiguous lane direction, we propagate the state forward by integrating the last observed speed $v$ and yaw rate $\omega$ for $T$ steps (with $\omega$ held constant; if $\omega \approx 0$, this reduces to straight-line motion). This yields a baseline trajectory $\{b_1,\dots,b_T\}$ in the lane-aligned frame. We also apply a global spatial scaling factor $S$ so that positions and residuals lie in a numerically stable range; at evaluation time, metrics are reported in meters by multiplying distances by $S$.

The model predicts a \emph{residual trajectory} relative to this baseline. For each future time step $t$, we define the ground-truth residual as
$\Delta p^{\mathrm{gt}}_t = p^{\mathrm{gt}}_t - b_t$, where $p^{\mathrm{gt}}_t$ is the true future position. The model therefore learns to output corrections to a nominal motion (e.g., steering adjustments or acceleration/braking), rather than absolute positions. This focuses learning on deviations from lane-following behavior and reduces dynamic range. Finally, we stack the 11 time-step vectors into an $11\times 9$ array $\mathbf{h}\in\mathbb{R}^{11\times 9}$ (rows $s_{-10},\dots,s_0$). Each training sample thus consists of $\mathbf{h}$ and a target residual
trajectory of length $T$.

\begin{algorithm}[t]
\caption{Quantum encoder (attention mechanism)}
\label{alg:encoder}
\DontPrintSemicolon
\KwIn{$\mathbf{h} \in \mathbb{R}^{11\times 9}$; $L_{\text{attn}}$; $\{\theta_i^{(\ell)}\}$}
\KwOut{$\mathbf{x}\in[-1,1]^9$}

$(s_{-10},\dots,s_0) \leftarrow \textsc{ReshapeHistory}(\mathbf{h})$\;
$\mathbf{q} \leftarrow s_0$\;
$\mathbf{k} \leftarrow \dfrac{1}{10}\sum_{i=-10}^{-1} s_i$\;
$\mathbf{v} \leftarrow s_0 - s_{-1}$\;
$(\mathbf{q},\mathbf{k},\mathbf{v}) \leftarrow \pi\tanh((\mathbf{q},\mathbf{k},\mathbf{v}))$\;

$n \leftarrow 9$\;
prepare $\ket{0}^{\otimes n}$\;

\For{$i \leftarrow 1$ \KwTo $n$}{
  apply $R_y(q_i)$ on qubit $i$\;
  apply $R_z(k_i)$ on qubit $i$\;
  apply $R_x(v_i)$ on qubit $i$\;
}

\For{$\ell \leftarrow 1$ \KwTo $L_{\text{attn}}$}{
  \For{$i \leftarrow 1$ \KwTo $n-1$}{
    apply $\mathrm{CX}(i\rightarrow i{+}1)$\;
    apply $R_z(\theta_i^{(\ell)})$ on qubit $i{+}1$\;
    apply $\mathrm{CX}(i\rightarrow i{+}1)$\;
  }
}

measure $Z_i$ on all qubits and set $\mathbf{x} \leftarrow (x_1,\dots,x_n)$\;
\Return{$\mathbf{x}$}\;
\end{algorithm}

\subsection{Quantum Encoder (Attention Mechanism)}

The first stage of our model is a quantum encoder that compresses the history into a latent representation of the SDV’s recent motion context. From $\mathbf{h}$ we derive three 9D vectors for an attention-like mechanism: the \emph{query} $\mathbf{q}\in\mathbb{R}^9$ is the current state $s_0$; the \emph{key} $\mathbf{k}\in\mathbb{R}^9$ is the mean of the 10 past states, summarizing overall history; and the \emph{value} $\mathbf{v}\in\mathbb{R}^9$ is the recent change $s_0 - s_{-1}$. These capture the current configuration, the long-term trend, and the short-term variation. The resulting quantum attention encoder that maps the history $\mathbf{h}$ to the context vector $\mathbf{x}$ is summarized in Algorithm~\ref{alg:encoder}.

We encode $(\mathbf{q},\mathbf{k},\mathbf{v})$ into a quantum state using $n=9$ qubits (one per feature). Starting from $\ket{0}^{\otimes 9}$, we apply a sequence of single-qubit rotations to inject the classical data. For qubit $i$, we apply $R_y(q_i)$, $R_z(k_i)$, and $R_x(v_i)$ in sequence, where $R_\alpha(\theta)$ is a rotation about axis $\alpha\in\{x,y,z\}$ by angle $\theta$. Before encoding, we apply a $\tanh$ normalization so that $q_i,k_i,v_i\in[-\pi,\pi]$, ensuring angles remain bounded. Denoting the encoding unitary by $U_{\text{enc}}$, the encoded state is
\begin{align}
\ket{\Psi_{\text{enc}}} &= U_{\text{enc}}(\mathbf{q},\mathbf{k},\mathbf{v})\ket{0}^{\otimes 9},\\
U_{\text{enc}}(\mathbf{q},\mathbf{k},\mathbf{v}) &= \prod_{i=1}^{9}\Big(R_x^{(i)}(v_i)\,R_z^{(i)}(k_i)\,R_y^{(i)}(q_i)\Big).
\end{align}

We then introduce entanglement to couple feature dimensions. For each neighboring pair $(i,i+1)$, $i=1,\dots,8$, we apply a controlled-NOT gate $\mathrm{CX}(i\rightarrow i+1)$, a trainable $R_z(\theta_i)$ on qubit $(i+1)$, and another $\mathrm{CX}(i\rightarrow i+1)$:
\begin{equation}
\ket{\Psi}\ \mapsto\ \mathrm{CX}_{i\to i+1}\,(I\otimes R_z(\theta_i))_{i+1}\,\mathrm{CX}_{i\to i+1}\,\ket{\Psi}.
\end{equation}
This implements a learned controlled phase rotation that entangles qubits $i$ and $i+1$, allowing adjacent features to influence each other. We refer to data encoding plus these entangling gates as one \emph{quantum attention layer}. We stack $L_{\text{attn}}=6$ such layers, each with its own parameters $\{\theta_1^{(\ell)},\dots,\theta_8^{(\ell)}\}$, enabling more complex interactions analogous to multi-layer or multi-head attention. The output of the final attention layer is an entangled multi-qubit state encoding the fused information of $\mathbf{q}$, $\mathbf{k}$, and $\mathbf{v}$.
To obtain a classical vector, we measure the expectation value of the Pauli-$Z$ operator on each qubit, yielding $\mathbf{x}\in\mathbb{R}^9$ with
\begin{equation}
x_i = \langle Z_i\rangle = \bra{\Psi_{\text{out}}}Z_i\ket{\Psi_{\text{out}}},\quad x_i\in[-1,1].
\end{equation}
This length-9 vector summarizes the motion context after attention-style mixing and serves as input to the quantum feedforward network.

\subsection{Quantum Feedforward Network}
\begin{algorithm}[t]
\caption{Quantum feedforward network}
\label{alg:ffn}
\DontPrintSemicolon
\KwIn{$\mathbf{x}^{(0)}\in[-1,1]^9$; $L_{\text{ff}}$; $\{\phi_i^{(\ell)},\psi_i^{(\ell)}\}$}
\KwOut{$\mathbf{z}\in[-1,1]^9$}

$n \leftarrow 9$\;

\For{$\ell \leftarrow 1$ \KwTo $L_{\text{ff}}$}{
  prepare $\ket{0}^{\otimes n}$\;

  \For{$i \leftarrow 1$ \KwTo $n$}{
    apply $R_y(x_i^{(\ell-1)})$ on qubit $i$\;
    apply $R_z(\phi_i^{(\ell)})$ on qubit $i$\;
    apply $R_y(\psi_i^{(\ell)})$ on qubit $i$\;
  }

  \For{$i \leftarrow 1$ \KwTo $n-1$}{
    apply $\mathrm{CX}(i\rightarrow i{+}1)$\;
  }
  apply $\mathrm{CX}(n\rightarrow 1)$\;

  measure $Z_i$ on all qubits and set
  $\mathbf{x}^{(\ell)} \leftarrow (x_1^{(\ell)},\dots,x_9^{(\ell)})$\;
}

$\mathbf{z} \leftarrow \tanh(\mathbf{x}^{(L_{\text{ff}})})$\;
\Return{$\mathbf{z}$}\;
\end{algorithm}

The encoder output $\mathbf{x}$ is passed to a deep quantum feedforward network that further refines the latent state. This network has $L_{\text{ff}}=64$ layers, each a parameterized circuit on the same $n=9$ qubits. The full encode–rotate–entangle–measure loop that refines $\mathbf{x}$ into the latent vector $\mathbf{z}$ is summarized in Algorithm~\ref{alg:ffn}.

In each feedforward layer, we reinitialize the qubits to $\ket{0}^{\otimes 9}$ and encode the current vector $\mathbf{x}$ via rotations: for each qubit $i$, we apply $R_y(x_i)$ (with $x_i$ scaled into an appropriate range if needed). We then apply two learned single-qubit rotations per qubit, $R_z(\phi_i)$ followed by $R_y(\psi_i)$. The angles $\{\phi_i,\psi_i\}_{i=1}^9$ are trainable parameters, playing a role analogous to weights in a fully-connected layer. Each feedforward layer thus has $2n$ parameters (18 when $n=9$), for about $1152$ parameters across all 64 layers.
To mix information across qubits, we introduce a ring entanglement pattern:
\begin{equation}
U_{\text{ring}}=\mathrm{CX}_{1\to2}\,\mathrm{CX}_{2\to3}\,\cdots\,\mathrm{CX}_{8\to9}\,\mathrm{CX}_{9\to1}.
\end{equation}
This cycle of CNOTs couples all qubits, so that each qubit’s state depends on the full vector $\mathbf{x}$. The entangling gates in the feedforward layers are fixed (non-trainable) and serve only to mix data.
At the end of each layer $\ell$, we measure Pauli-$Z$ expectations to
obtain a new 9D vector \begin{equation}
\mathbf{x}^{(\ell)}=\big(\langle Z_1 \rangle^{(\ell)},\dots,\langle Z_9 \rangle^{(\ell)}\big),
\end{equation}
which forms the input to the next layer. Repeating this encode-rotate-entangle-measure loop for $L_{\text{ff}}{=}64$ layers yields a final latent vector $\mathbf{z}\in\mathbb{R}^9$.
We apply an elementwise $\tanh$ to $\mathbf{z}$ to keep components in $[-1,1]$. The vector $\mathbf{z}$ is a compact representation of the scenario, distilled from the SDV’s past trajectory and context, and is used by the decoder to generate multi-modal futures.

\subsection{Quantum Decoder for Multi-Modal Prediction}
\begin{algorithm}[t]
\caption{Quantum decoder for multi-modal prediction}
\label{alg:decoder}
\DontPrintSemicolon
\KwIn{$\mathbf{z}\in[-1,1]^9$; $b_{1:T}$; $T{=}20$; $M$; $B$; $S_r$; $\{\gamma_i\}$}
\KwOut{$\{p^{(m)}_{1:T}\}_{m=1}^M$; $\{\pi^{(m)}\}_{m=1}^M$}

$n \leftarrow 9$\;
prepare $\ket{0}^{\otimes n}$\;

\For{$i \leftarrow 1$ \KwTo $n$}{
  apply $R_y(z_i)$ on qubit $i$\;
  apply $R_z(\gamma_i)$ on qubit $i$\;
}

\For{$i \leftarrow 1$ \KwTo $n-1$}{
  apply $\mathrm{CX}(i\rightarrow i{+}1)$\;
  apply $R_y(\gamma_i)$ on qubit $i{+}1$\;
  apply $\mathrm{CX}(i\rightarrow i{+}1)$\;
}

$\alpha \leftarrow \textsc{StateVector}() \in \mathbb{C}^{2^n}$\;

\For{$m \leftarrow 1$ \KwTo $M$}{
  $\Delta\phi_m \leftarrow \dfrac{(m+1)\pi}{M}$\;
  $\alpha^{(m)} \leftarrow \textsc{ApplyPhase}(\alpha,\Delta\phi_m)$\;

  \For{$t \leftarrow 1$ \KwTo $T$}{
    $\Delta x_t^{(m)} \leftarrow \displaystyle\sum_{j=1}^{B}
      \Re(\alpha^{(m)}_j)\cos\!\big(\tfrac{j\pi t}{T+1}\big)$\;
    $\Delta y_t^{(m)} \leftarrow \displaystyle\sum_{j=1}^{B}
      \Im(\alpha^{(m)}_j)\sin\!\big(\tfrac{j\pi t}{T+1}\big)$\;
    $\Delta p_t^{(m)} \leftarrow S_r(\Delta x_t^{(m)},\Delta y_t^{(m)})$\;
    $p_t^{(m)} \leftarrow b_t + \Delta p_t^{(m)}$\;
  }
}

$\mathbf{z}' \leftarrow \textsc{PadToLength}(\mathbf{z},M)$\;
$\hat{\mathbf{z}} \leftarrow \mathcal{F}(\mathbf{z}')$\;

\For{$m \leftarrow 1$ \KwTo $M$}{
  $c_m \leftarrow |\hat{z}_m|$\;
}
\For{$m \leftarrow 1$ \KwTo $M$}{
  $\pi^{(m)} \leftarrow c_m / \sum_{k=1}^M c_k$\;
}

\Return{$\{p^{(m)}_{1:T}\}_{m=1}^M$, $\{\pi^{(m)}\}_{m=1}^M$}\;
\end{algorithm}

The latent vector $\mathbf{z}$ is fed into a quantum decoder that outputs $M$ future trajectory hypotheses and a confidence for each. We use $M=16$ modes. The decoder is designed so that a single quantum circuit, together with mode-specific phase shifts, produces all $M$ modes via quantum superposition and interference. Algorithm~\ref{alg:decoder} details the quantum decoder that maps the latent $\mathbf{z}$ to the set of trajectories $\{p^{(m)}_{1:T}\}_{m=1}^M$ and their confidences $\{\pi^{(m)}\}_{m=1}^M$.

The decoder uses another parameterized $n$-qubit circuit. We encode $\mathbf{z}$ by applying $R_y(z_i)$ to qubit $i$. We then apply a trainable $R_z(\gamma_i)$ on each qubit, followed by a sequential ``ladder'' entanglement pattern: for each $i=1,\dots,8$, we apply $\mathrm{CX}(i\rightarrow i+1)$, then $R_y(\gamma_i)$ on qubit $(i+1)$, then another $\mathrm{CX}(i\rightarrow i+1)$. Using the same parameter $\gamma_i$ for both $R_z$ and the ladder coupling allows the circuit to distribute latent information across the amplitudes of different computational basis states. The resulting state
$\ket{\Psi(\mathbf{z})}$ is a superposition over all $2^n$ basis states.

To generate mode-specific trajectories, we exploit global phase shifts. For each mode $m\in\{1,\dots,M\}$, we start from $\ket{\Psi(\mathbf{z})}$ and apply $R_z(\Delta\phi_m)$ to every qubit, where $\Delta\phi_m = \frac{(m+1)\pi}{M}$. This changes the relative phases but not the basis probabilities, producing a new state $\ket{\Psi^{(m)}}$. We then extract the complex amplitude vector $\alpha^{(m)}\in\mathbb{C}^{2^n}$, where $\alpha^{(m)}_j = \langle j \vert \Psi^{(m)}\rangle$ for $j=0,\dots,2^n-1$.

We map $\alpha^{(m)}$ deterministically to a 2D trajectory using a truncated Fourier series. For horizon $T=20$, we choose truncation order $B=8$. Ignoring the all-zero basis state ($j=0$), we use the next $B$ amplitudes as Fourier coefficients. The real parts of these $B$ amplitudes weight cosine basis functions, and the imaginary parts weight sine basis functions. For mode $m$, the residual trajectory $\Delta p^{(m)}_t = (\Delta x^{(m)}_t,\Delta y^{(m)}_t)$ at time $t$ is
\begin{align}
\Delta x^{(m)}_{t} &= \sum_{j=1}^{B} \Re\!\big(\alpha^{(m)}_j\big)\, \cos\!\Big(\frac{j\pi t}{T+1}\Big),\\
\Delta y^{(m)}_{t} &= \sum_{j=1}^{B} \Im\!\big(\alpha^{(m)}_j\big)\, \sin\!\Big(\frac{j\pi t}{T+1}\Big),
\end{align}
for $t=1,\dots,T$. We then apply a scale factor $S_r=1.5$ to match the magnitude of residuals to the normalized data,
\begin{equation}
\Delta p_t^{(m)} = S_r
\begin{pmatrix}
\Delta x_t^{(m)}\\[2pt]
\Delta y_t^{(m)}
\end{pmatrix},
\end{equation}
and add the baseline trajectory to obtain absolute predictions:
\begin{equation}
p_t^{(m)} = b_t + \Delta p_t^{(m)}.
\end{equation}
If desired, positions can be mapped back to world coordinates by inverting the lane-aligned transform and scaling.

This decoder yields $M$ distinct trajectories $\{p^{(m)}_{1:T}\}_{m=1}^M$ from a single latent $\mathbf{z}$ and a shared set of parameters. Different phase offsets $\Delta\phi_m$ produce diverse interference patterns in the Fourier coefficients and thus diverse, yet smooth, trajectories (e.g., straight vs.\ turn). Crucially, multi-modality arises from phase variation rather than $M$ separate networks, so all modes are generated in one forward pass.

The decoder also outputs a confidence for each mode. We compute these by analyzing the frequency content of $\mathbf{z}$. Specifically, we take the discrete Fourier transform (DFT) of a zero-padded latent vector $\mathbf{z}'\in\mathbb{R}^{16}$ and use the magnitudes of the $M$ Fourier coefficients as raw scores:
\begin{equation}
c_m = \big|(\mathcal{F}(\mathbf{z}'))_m\big|,\quad m=1,\dots,M.
\end{equation}
Normalizing these yields a confidence distribution $(\pi^{(1)},\dots,\pi^{(M)})$:
\begin{equation}
\pi^{(m)} = \frac{c_m}{\sum_{k=1}^{M} c_k}.
\end{equation}
Although this confidence model is not learned, it provides a geometry-aware ranking: modes that align more strongly with dominant latent frequencies receive higher scores. We use these confidences for ranking-based metrics such as recall and precision.

\subsection{Loss Function and Training Strategy}
\begin{algorithm}[t]
\caption{SPSA training for quantum multi-modal trajectory model}
\label{alg:training}
\DontPrintSemicolon
\KwIn{triplets $(\mathbf{h}, b_{1:T}, \Delta p^{\mathrm{gt}}_{1:T})$; number of epochs; batches per epoch; batch size; $a,c,A,\alpha,\gamma$; $\lambda$}
\KwOut{parameters $\boldsymbol{\theta}$}

initialize $\boldsymbol{\theta}$\;

\For{epoch $=1$ \KwTo number of epochs}{
  $k \leftarrow 0$\;
  partition and shuffle triplets into batches of given batch size\;

  \ForEach{batch}{
    \ForEach{$(\mathbf{h}, b_{1:T}, \Delta p^{\mathrm{gt}}_{1:T})$ in batch}{
      $k \leftarrow k+1$\;
      $\Delta \leftarrow \textsc{Rand}\{\pm 1\}^{|\boldsymbol{\theta}|}$\;
      $a_k \leftarrow a(A{+}k)^{-\alpha}$\;
      $c_k \leftarrow c\,k^{-\gamma}$\;

      $\boldsymbol{\theta}^{+} \leftarrow \boldsymbol{\theta} + c_k \Delta$\;
      $\boldsymbol{\theta}^{-} \leftarrow \boldsymbol{\theta} - c_k \Delta$\;

      $\{p^{(m,+)}_{1:T}\}_{m=1}^M \leftarrow \textsc{Forward}(\mathbf{h}, b_{1:T}, \boldsymbol{\theta}^{+})$\;
      $L^{+} \leftarrow \textsc{Loss}(\{p^{(m,+)}_{1:T}\}_{m=1}^M, b_{1:T}, \Delta p^{\mathrm{gt}}_{1:T}, \lambda)$\;

      $\{p^{(m,-)}_{1:T}\}_{m=1}^M \leftarrow \textsc{Forward}(\mathbf{h}, b_{1:T}, \boldsymbol{\theta}^{-})$\;
      $L^{-} \leftarrow \textsc{Loss}(\{p^{(m,-)}_{1:T}\}_{m=1}^M, b_{1:T}, \Delta p^{\mathrm{gt}}_{1:T}, \lambda)$\;

      $g \leftarrow \dfrac{L^{+} - L^{-}}{2 c_k}\,\Delta$\;
      $\boldsymbol{\theta} \leftarrow \boldsymbol{\theta} - a_k g$\;
    }
  }
}

\Return{$\boldsymbol{\theta}$}\;
\end{algorithm}

We train the model by comparing the predicted multi-modal trajectories to the ground truth and minimizing a regression loss that accounts for multiple modes. Given the $M$ predicted trajectories $\{p^{(m)}_{1:T}\}_{m=1}^M$ for a sample and the ground-truth trajectory $\{p^{\text{gt}}_t\}_{t=1}^T$, we define
\begin{equation}
\begin{split}
\mathcal{L} := \min_{m=1,\dots,M}\frac{1}{T}\sum_{t=1}^{T} \big\|p^{(m)}_t-p^{\mathrm{gt}}_t\big\|^2 \\
\qquad + \lambda\,\frac{1}{MT}\sum_{m=1}^M\sum_{t=1}^T \big\|\Delta p^{(m)}_t\big\|^2
\end{split}
\end{equation}
where $\|\cdot\|$ is the Euclidean norm in $\mathbb{R}^2$ and $\lambda$ is a small weighting coefficient. The first term is the mean squared error (MSE) of the best-predicting mode, encouraging at least one trajectory to match the true outcome. The second term penalizes large residuals and discourages overly oscillatory predictions; we use $\lambda = 10^{-4}$ so that this acts as a mild smoothness regularizer. The loss is computed in normalized lane-frame coordinates; for reporting in meters, we rescale distances by $S$ at evaluation time. The SPSA-based training loop used to minimize this loss and update all circuit parameters $\boldsymbol{\theta}$ is summarized in Algorithm~\ref{alg:training}.

Because the model is a hybrid quantum–classical system with non-analytic components (e.g., the discrete Fourier mapping and $\min$ operator), standard backpropagation through all layers is nontrivial. We therefore use a gradient-free optimizer, SPSA to train all circuit parameters. SPSA is well-suited to high-dimensional optimization with noisy function evaluations. At each iteration, it perturbs the full parameter vector $\boldsymbol{\theta}$ (all trainable angles in the encoder, feedforward stack, and decoder, $\sim\!1200$ parameters) in two opposite random directions and evaluates the loss at $\boldsymbol{\theta}+\Delta$ and $\boldsymbol{\theta}-\Delta$. The gradient is estimated as
\begin{equation}
\nabla L(\boldsymbol{\theta}) \;\approx\;
\frac{L(\boldsymbol{\theta}+c_k\,\Delta)-L(\boldsymbol{\theta}-c_k\,\Delta)}{2\,c_k}\,\Delta
\end{equation}
where $\Delta$ has i.i.d.\ components in $\{\pm1\}$ and $c_k$ is a perturbation
scale at iteration $k$. Crucially, SPSA requires only two loss evaluations per iteration, regardless of the number of parameters.

We use decaying step sizes $a_k$ and perturbation scales $c_k$,
\begin{equation}
a_k = a\,(A+k)^{\alpha}, \qquad c_k = c\,k^{\gamma},
\end{equation}
with hyperparameters $A=80$, $a=0.05$, $c=0.1$, $\alpha=0.602$, and $\gamma=0.101$, following standard SPSA practice. We further reduce variance by averaging the gradient estimate over two independent perturbation draws per iteration.
\begin{figure*}[t]
  \centering

  \begin{subfigure}[t]{0.30\textwidth}
    \centering
    \includegraphics[width=0.90\linewidth]{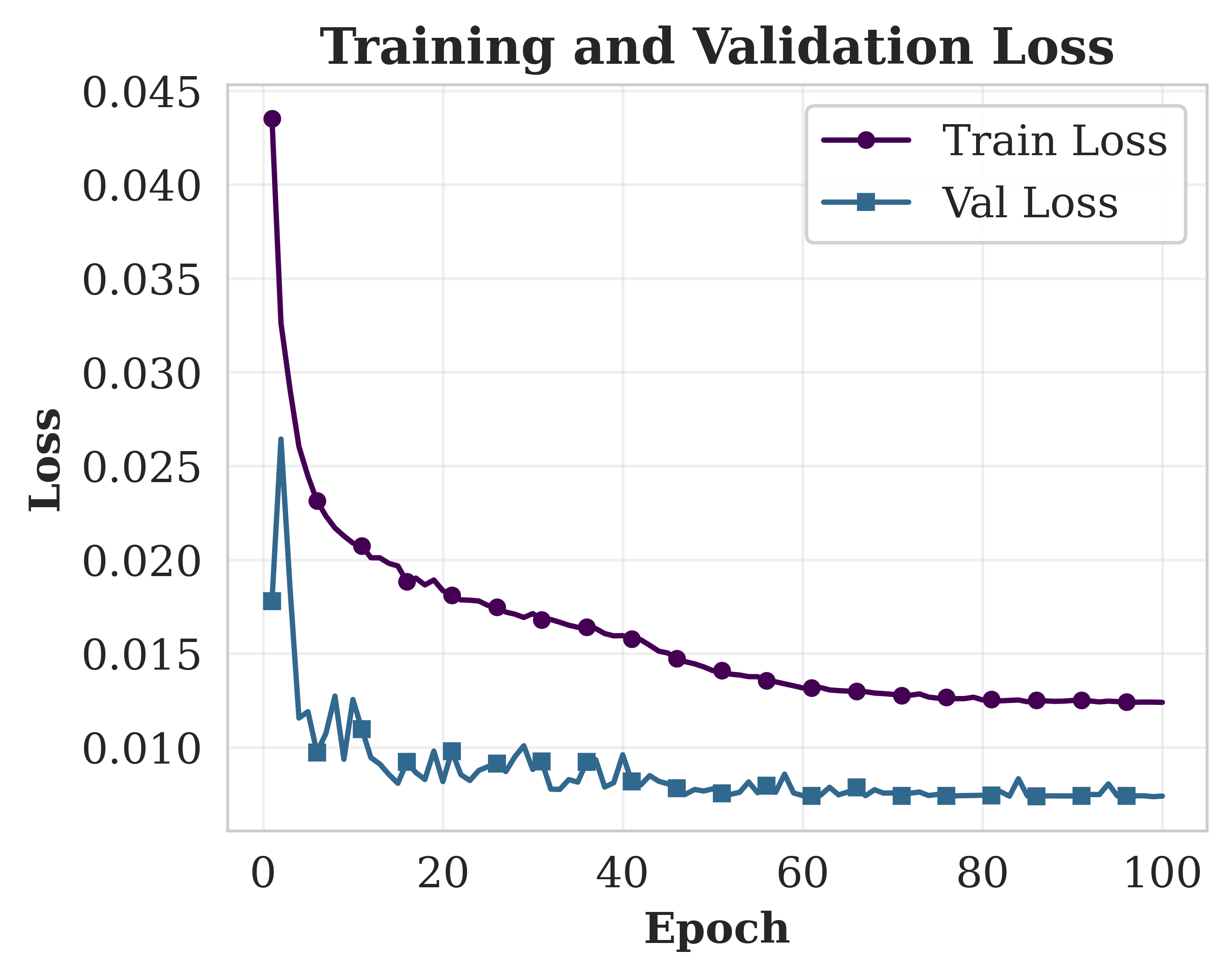}
    \caption{Global loss behaviour over 100 epochs.}
    \label{fig:global-loss}
  \end{subfigure}\hfill
  \begin{subfigure}[t]{0.30\textwidth}
    \centering
    \includegraphics[width=0.90\linewidth]{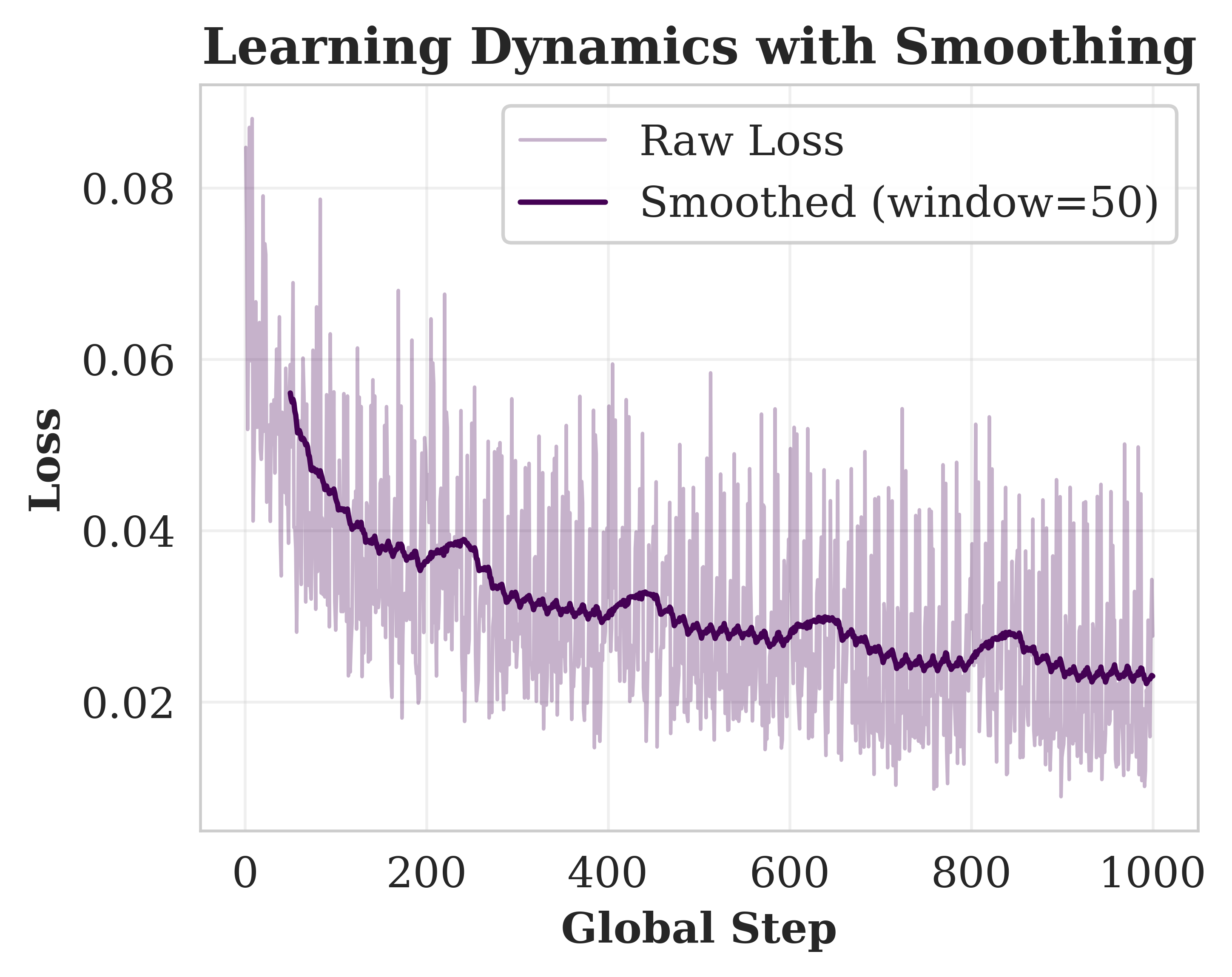}
    \caption{Loss (raw \& smoothed).}
    \label{fig:signals-loss-dynamics}
  \end{subfigure}\hfill
  \begin{subfigure}[t]{0.30\textwidth}
    \centering
    \includegraphics[width=0.90\linewidth]{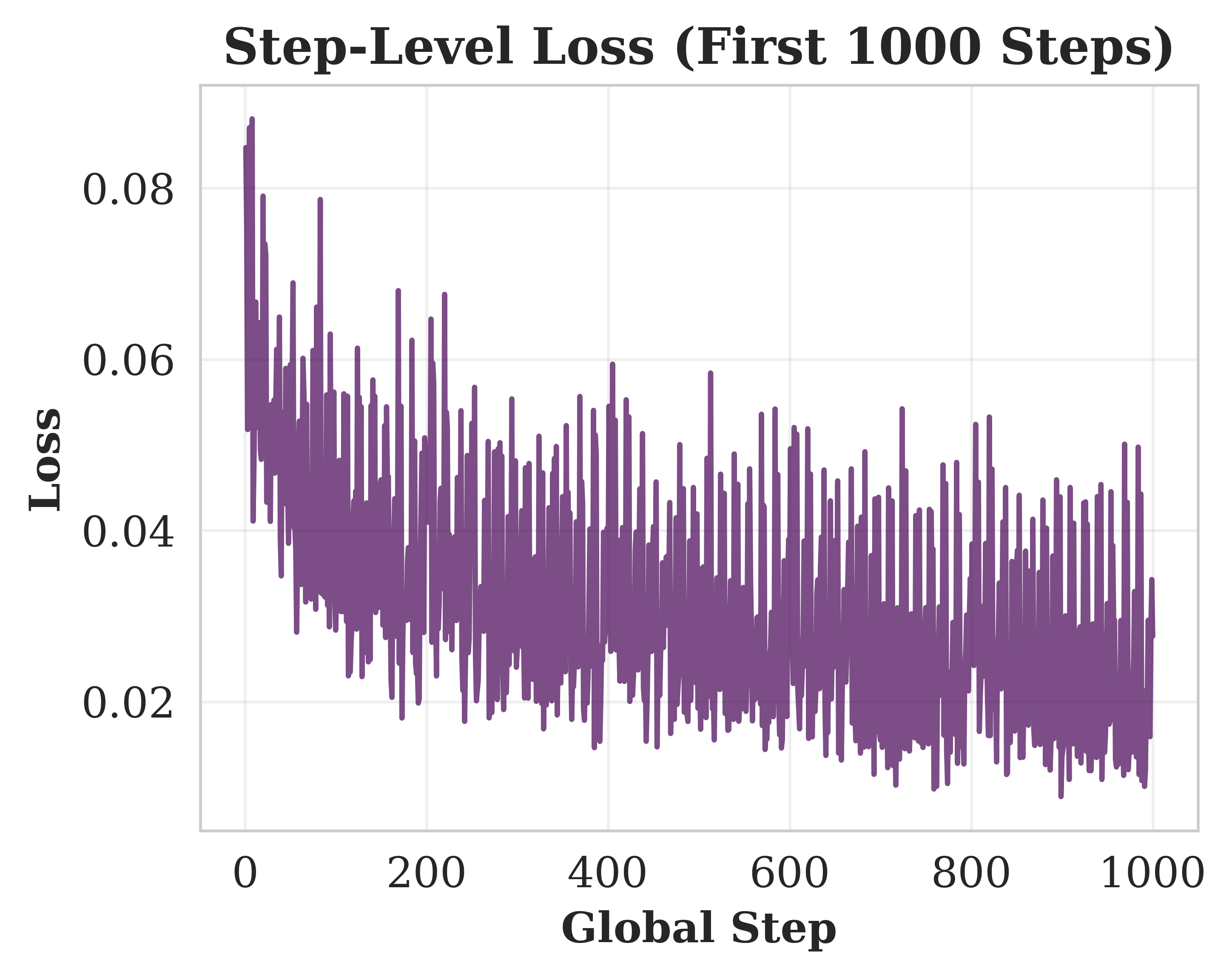}
    \caption{Per-step loss trace.}
    \label{fig:signals-step-loss}
  \end{subfigure}

  \caption{Global and step-level training signals. \textbf{(a)} Training and validation losses decrease smoothly and track closely, indicating good generalization under residual prediction and bounded-angle normalization. \textbf{(b)} Raw and smoothed loss show steady improvement with gentle undulations aligned with SPSA schedule resets. \textbf{(c)} Step-wise loss traces confirm the same trend at finer granularity.}
  \label{fig:training-signals-overview}
\end{figure*}
Training runs for 100 epochs. Each epoch consists of 200 randomly sampled batches from the training set, with batch size 32. Within a batch, we apply SPSA updates per sample (an online training style) rather than aggregating gradients across the batch. We reset the SPSA iteration counter at the start of each epoch, effectively restarting the learning-rate schedule and preventing overly rapid decay. After each epoch, we evaluate the model on a held-out validation subset using the same preprocessing and decoder to obtain all $M$ trajectories, which are then compared to ground truth. All circuit parameters are initialized with small random values drawn from $\mathcal{N}(0,0.05^2)$, and a fixed random seed is used for initialization and data shuffling to ensure reproducibility. We also save parameter checkpoints $\boldsymbol{\theta}$ periodically for potential early stopping and post-hoc analysis.

\begin{figure*}[t!]
  \centering

  \begin{subfigure}[t]{0.30\textwidth}
    \centering
    \includegraphics[width=0.90\linewidth]{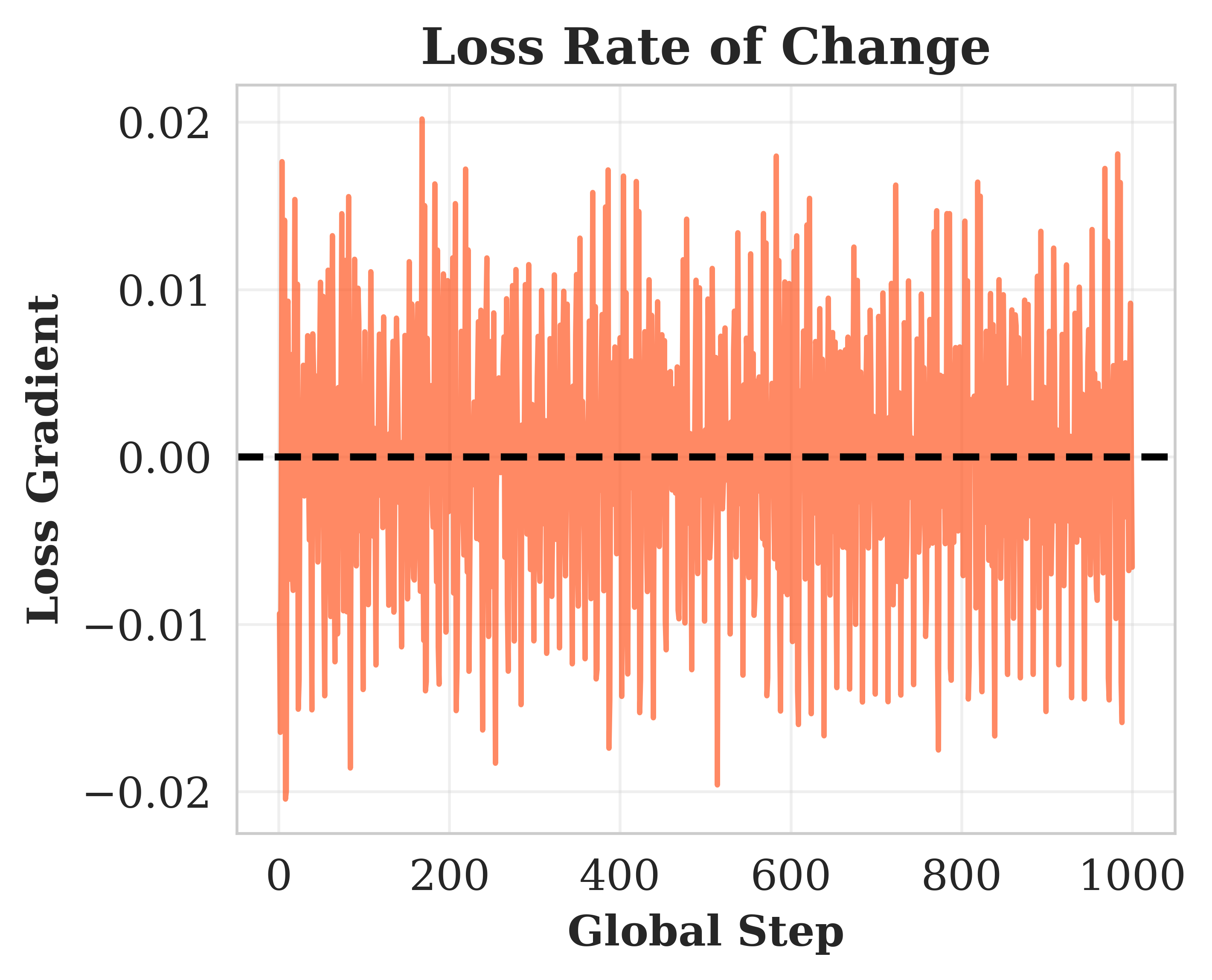}
    \caption{Finite-difference loss rate.}
    \label{fig:signals-loss-rate}
  \end{subfigure}\hfill
  \begin{subfigure}[t]{0.30\textwidth}
    \centering
    \includegraphics[width=0.90\linewidth]{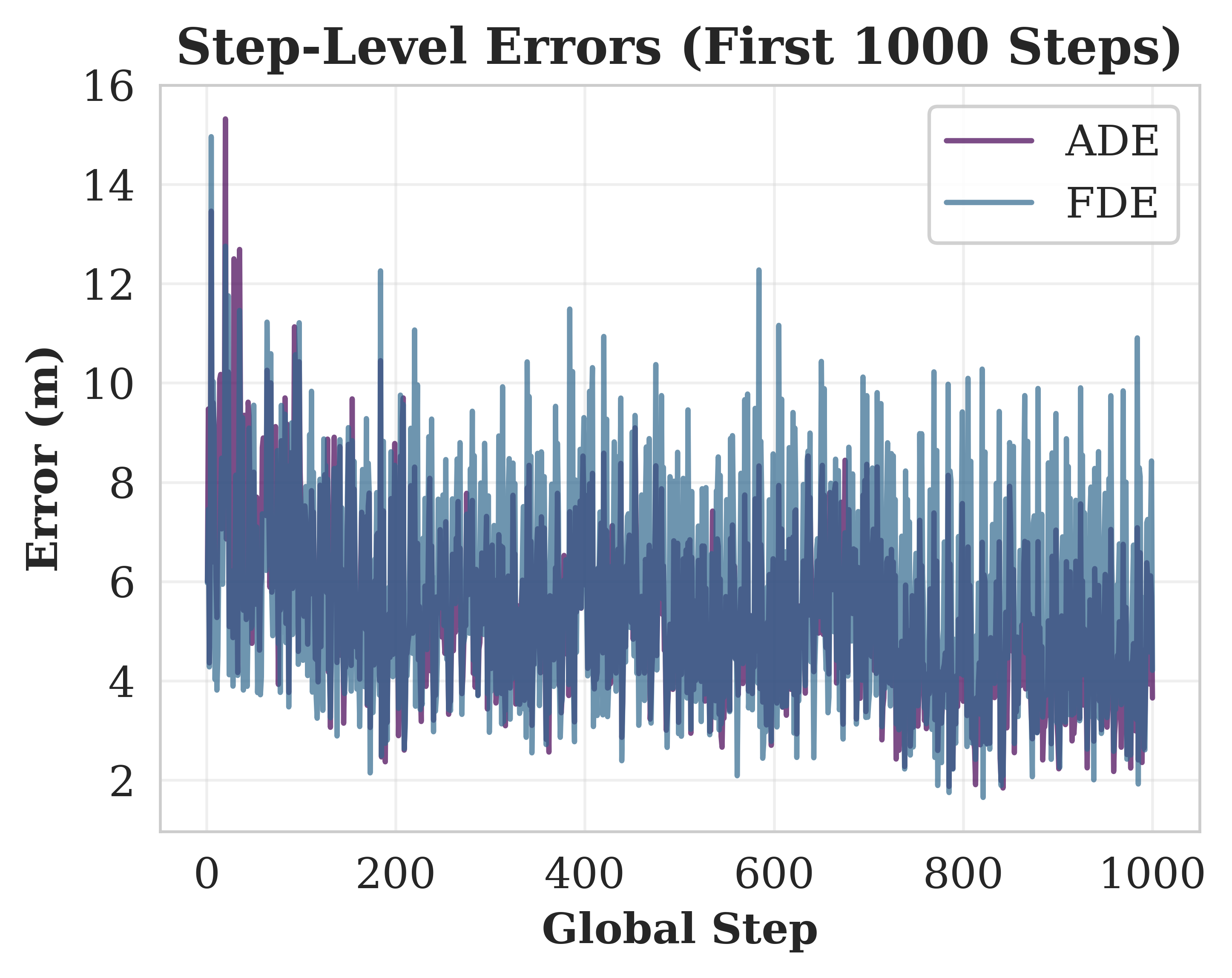}
    \caption{Step-level ADE/FDE.}
    \label{fig:signals-step-errors}
  \end{subfigure}\hfill
  \begin{subfigure}[t]{0.30\textwidth}
    \centering
    \includegraphics[width=0.90\linewidth]{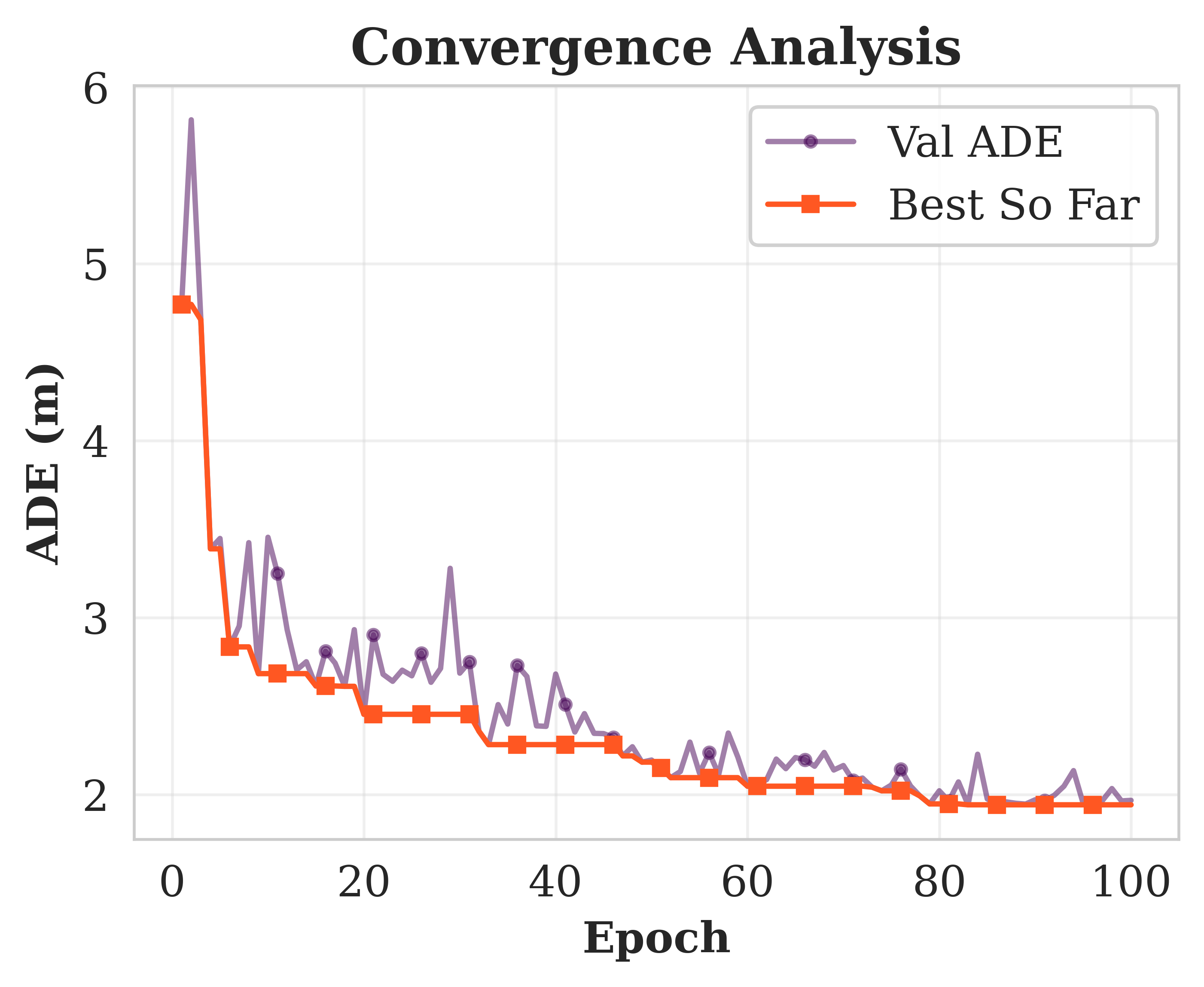}
    \caption{Best-so-far validation ADE.}
    \label{fig:conv-035}
  \end{subfigure}

  \caption{Fine-grained learning signals and discrete convergence steps. \textbf{(a)} Finite-difference loss rate oscillates around a negative mean; sign flips reflect mode switches under the $\min$-over-modes objective while the global trend decreases. \textbf{(b)} Step-level ADE/FDE amplitudes contract over training, confining volatility to a shrinking subset of scenes. \textbf{(c)} Best-so-far validation ADE forms a staircase with drops at epochs $\sim$6, 18, 35, 52, 74, and 83, consistent with phase realignments between the shallow attention/decoder and the Fourier readout.}
  \label{fig:fine-grained-signals}
\end{figure*}

\section{Experiments and Results}
\subsection{Evaluation Setup}


\paragraph{Prediction Horizon and Output Format.}
For each scenario the model produces $M{=}16$ trajectory hypotheses, each covering a 2.0\,s horizon at 0.1\,s resolution, together with confidence scores $\pi^{(m)}$. We interpret these as $M$ plausible futures for the SDV: the highest-confidence trajectory can be treated as the primary prediction, while the remaining modes capture uncertainty and multi-modality in complex scenes.

\paragraph{Validation Procedure.}
During validation, each scenario’s 1.1\,s history is fed through the model to obtain the $M{=}16$ predicted trajectories and confidences; displacement and ranking metrics are then computed per scenario and aggregated. Errors are evaluated in the local lane-aligned frame and converted to meters for reporting. We also evaluate a simple baseline (lane-following or CTRV extrapolation from preprocessing) by comparing its trajectory $\{b_t\}$ (zero residual) to the ground truth, yielding baseline ADE/FDE. In practice, the model’s minADE and minFDE are significantly lower than the baseline’s errors, indicating that the quantum model learns meaningful residual corrections to the nominal path. For all metrics we use a large, representative subset of the validation set, 50 batches of 32 scenarios (1{,}600 scenarios), due to simulation constraints, and report averages over these scenes. We repeat evaluation with different random seeds and observe minimal variation, suggesting stable performance and reliable estimates. The reported quantities (minADE, minFDE, Recall@K, mAP) can therefore be regarded as robust indicators of the model’s accuracy on the validation split.

\subsection{Training dynamics and convergence}

\begin{figure*}[t]
  \centering

  \begin{subfigure}[t]{0.3\textwidth}
    \centering
    \includegraphics[width=0.90\linewidth]{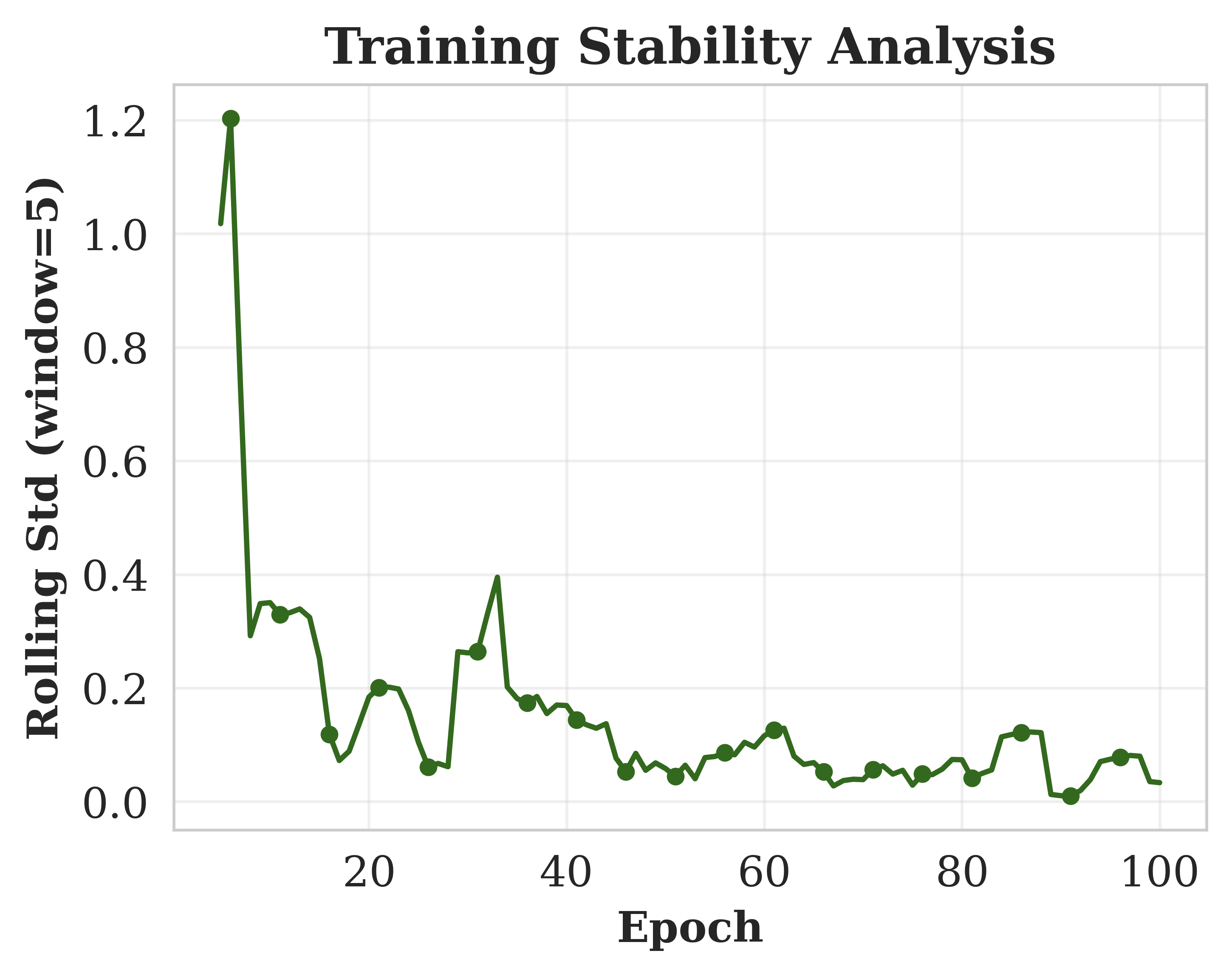}
    \caption{Rolling stdev of ADE collapses.}
    \label{fig:conv-034}
  \end{subfigure}\hfill
  \begin{subfigure}[t]{0.3\textwidth}
    \centering
    \includegraphics[width=0.90\linewidth]{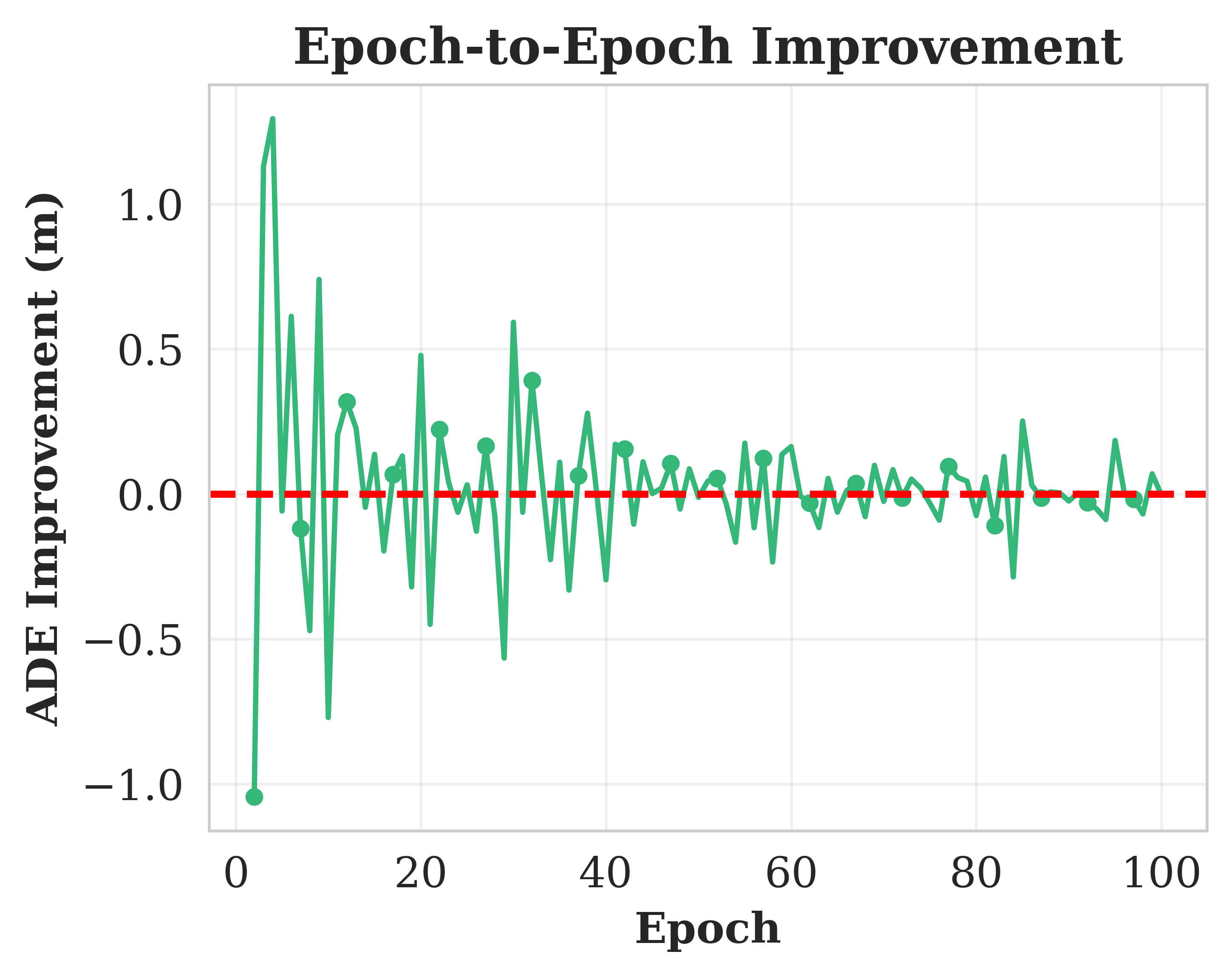}
    \caption{Epoch-to-epoch improvement rate tapers.}
    \label{fig:conv-032}
  \end{subfigure}\hfill
  \begin{subfigure}[t]{0.3\textwidth}
    \centering
    \includegraphics[width=0.90\linewidth]{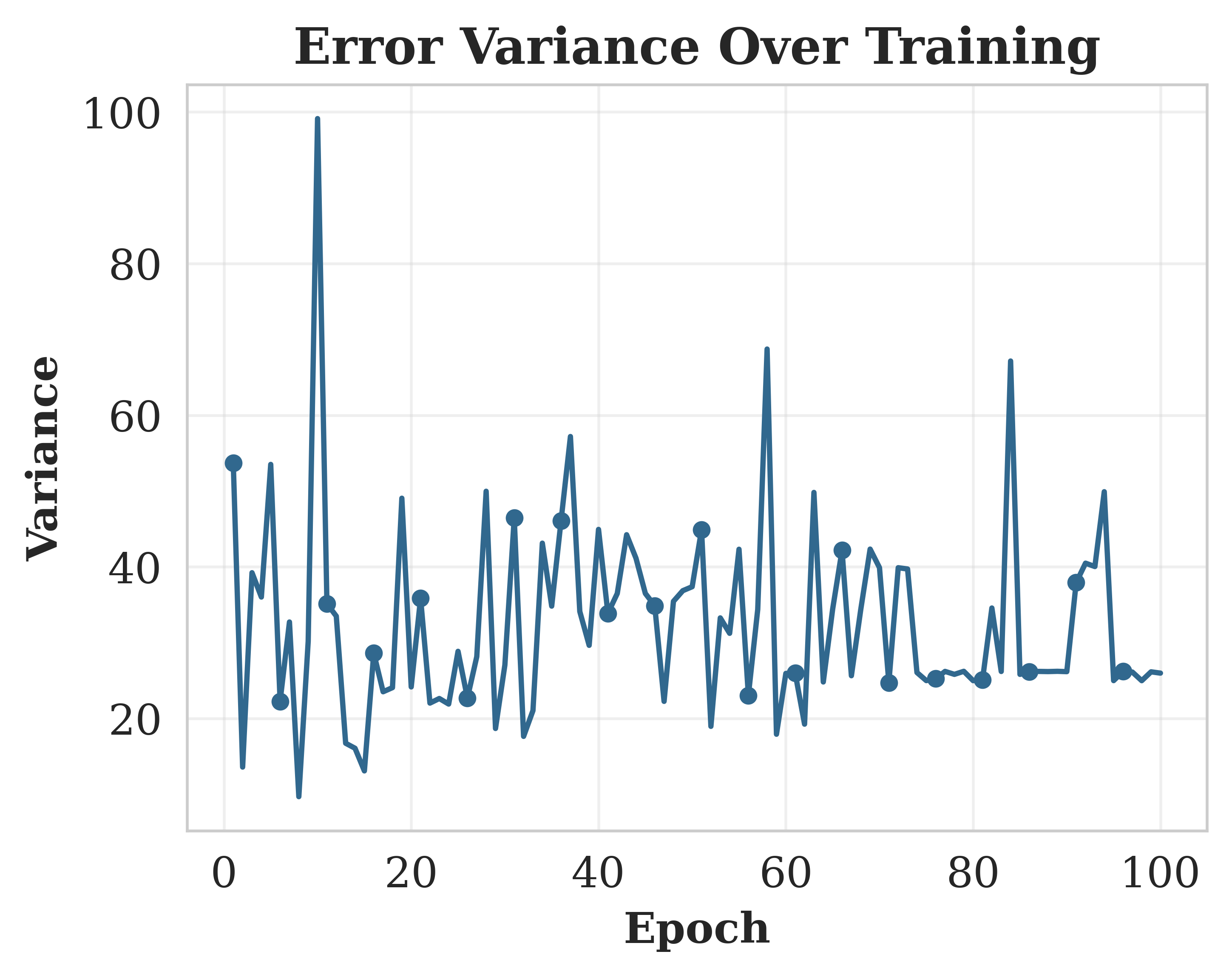}
    \caption{Variance settles to a tight band.}
    \label{fig:conv-031}
  \end{subfigure}

  \caption{Convergence and stability diagnostics under SPSA. \textbf{(a)} Rolling ADE standard deviation drops by roughly an order of magnitude from early to late training. \textbf{(b)} Epoch-to-epoch improvement rate tapers after $\sim$70 epochs, indicating approach to a stable optimum. \textbf{(c)} Error variance settles into a narrow band. Together these trends indicate stable optimization and predictable convergence for the shallow 9-qubit circuits.}
  \label{fig:convergence-stability}
\end{figure*}
\begin{figure*}[t!]
  \centering

  \begin{subfigure}[t]{0.3\textwidth}
    \centering
    \includegraphics[width=0.9\linewidth]{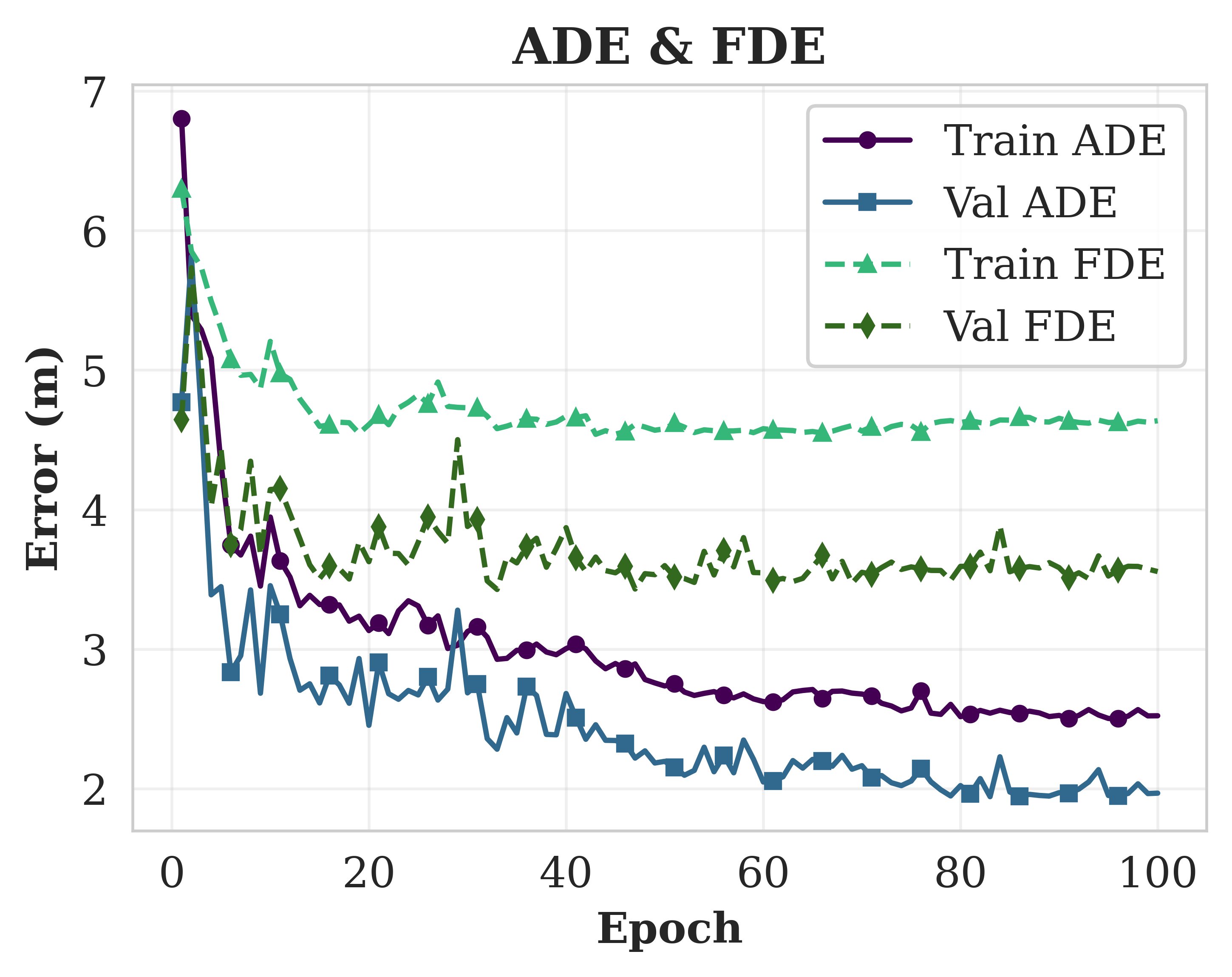}
    \caption{ADE/FDE across epochs.}
    \label{fig:acc-ade-fde}
  \end{subfigure}\hfill
  \begin{subfigure}[t]{0.3\textwidth}
    \centering
    \includegraphics[width=0.9\linewidth]{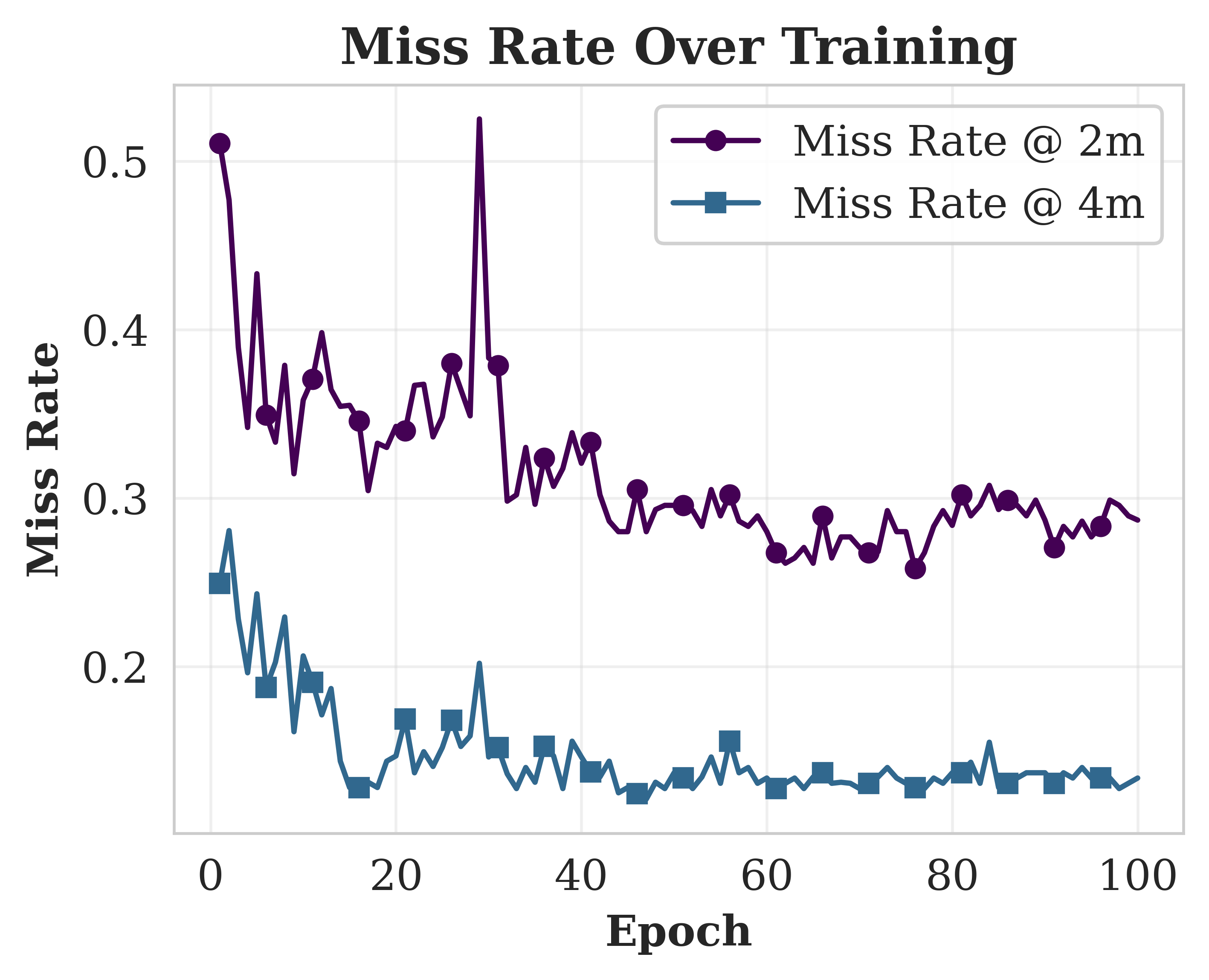}
    \caption{Miss@\SI{2}{m} and Miss@\SI{4}{m}}
    \label{fig:acc-miss}
  \end{subfigure}\hfill
  \begin{subfigure}[t]{0.3\textwidth}
    \centering
    \includegraphics[width=0.9\linewidth]{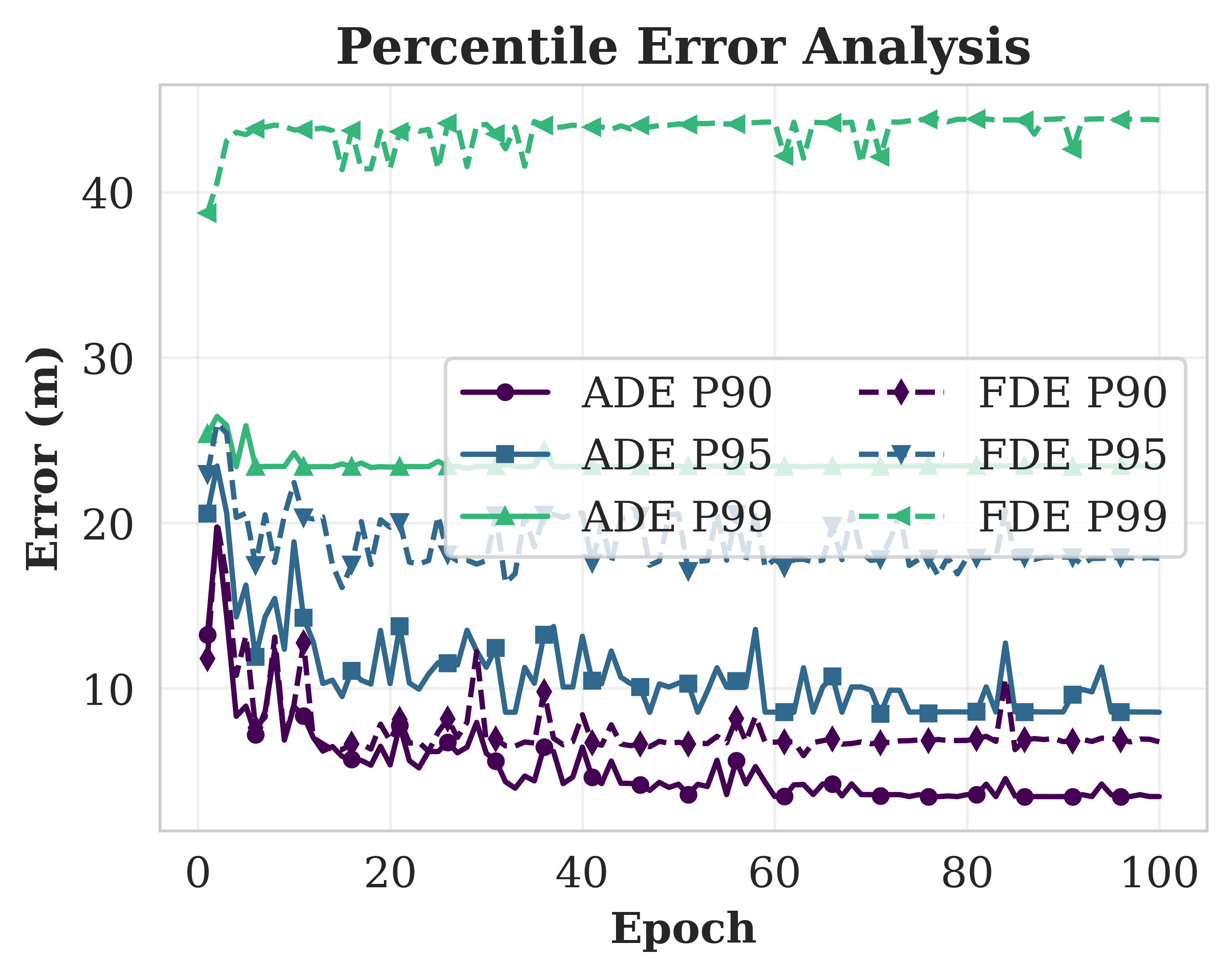}
    \caption{Percentile ADE (P50–P99).}
    \label{fig:acc-percentiles}
  \end{subfigure}

  \vspace{0.8em}

  \begin{subfigure}[t]{0.32\textwidth}
    \centering
    \includegraphics[width=\linewidth]{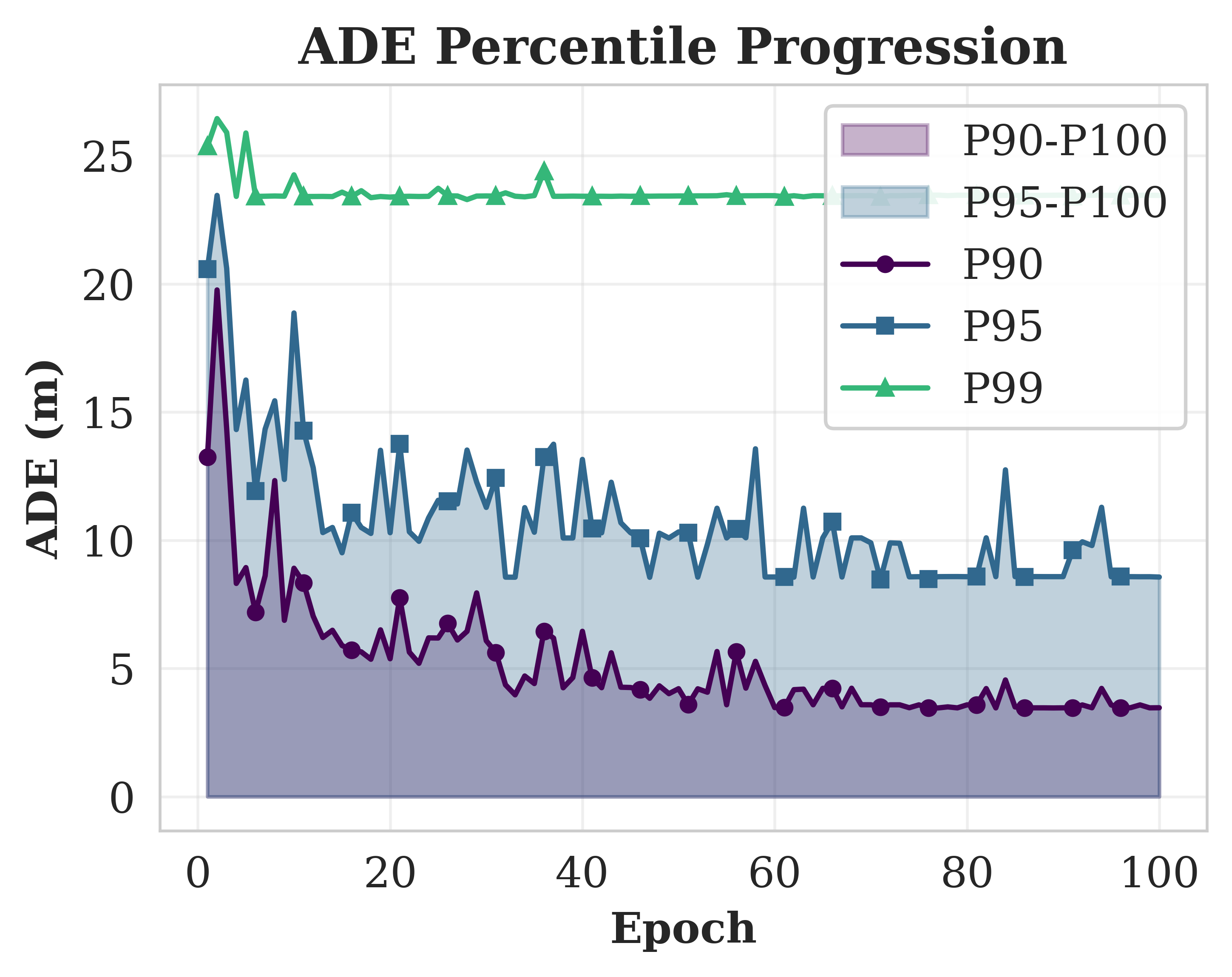}
    \caption{Progression with shaded P90–P100 and P95–P100.}
    \label{fig:acc-percentile-bands}
  \end{subfigure}\hfill
  \begin{subfigure}[t]{0.32\textwidth}
    \centering
    \includegraphics[width=\linewidth]{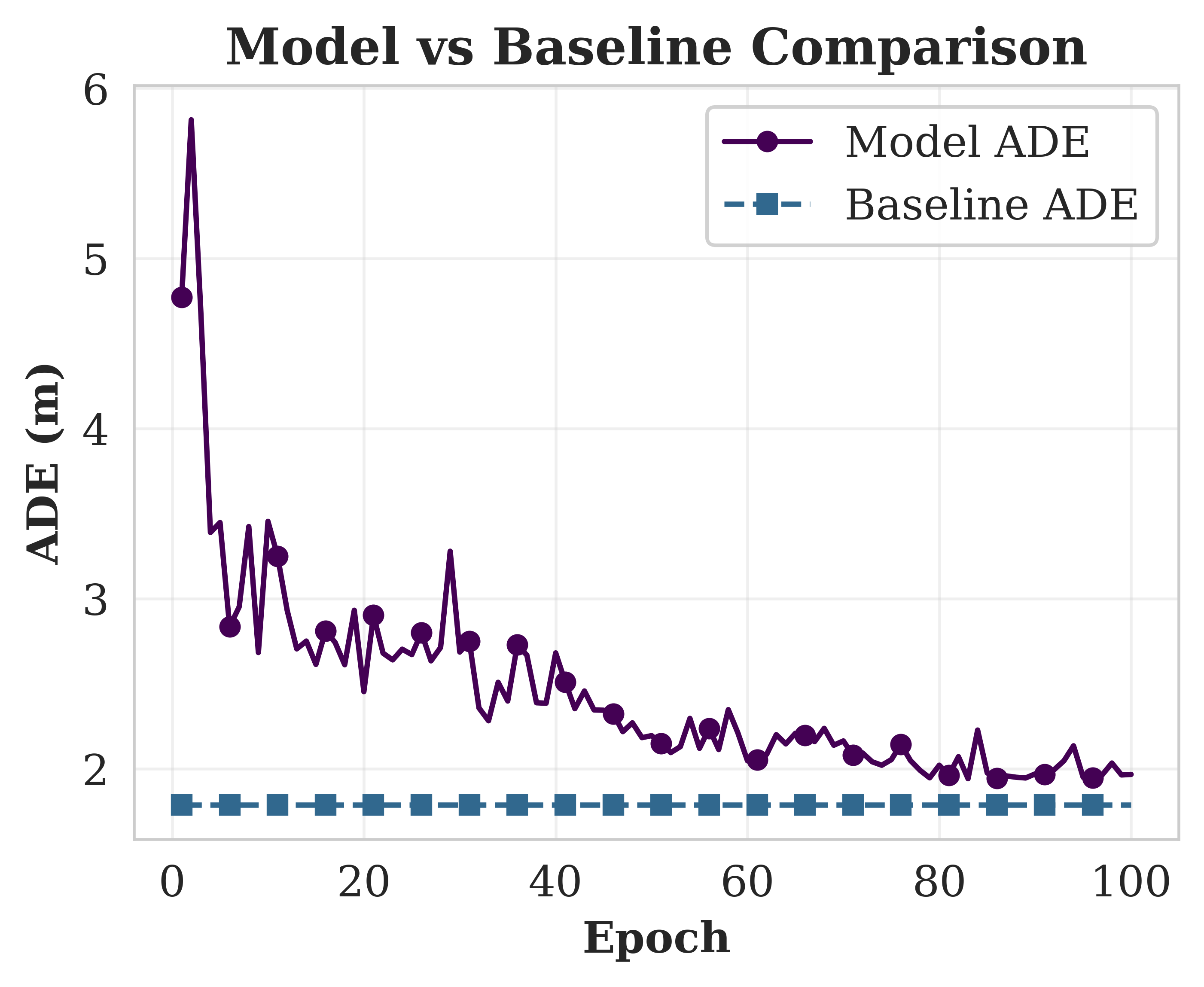}
    \caption{ADE vs. lane/CTRV baseline.}
    \label{fig:acc-kinematic}
  \end{subfigure}\hfill
  \begin{subfigure}[t]{0.32\textwidth}
    \centering
    \includegraphics[width=\linewidth]{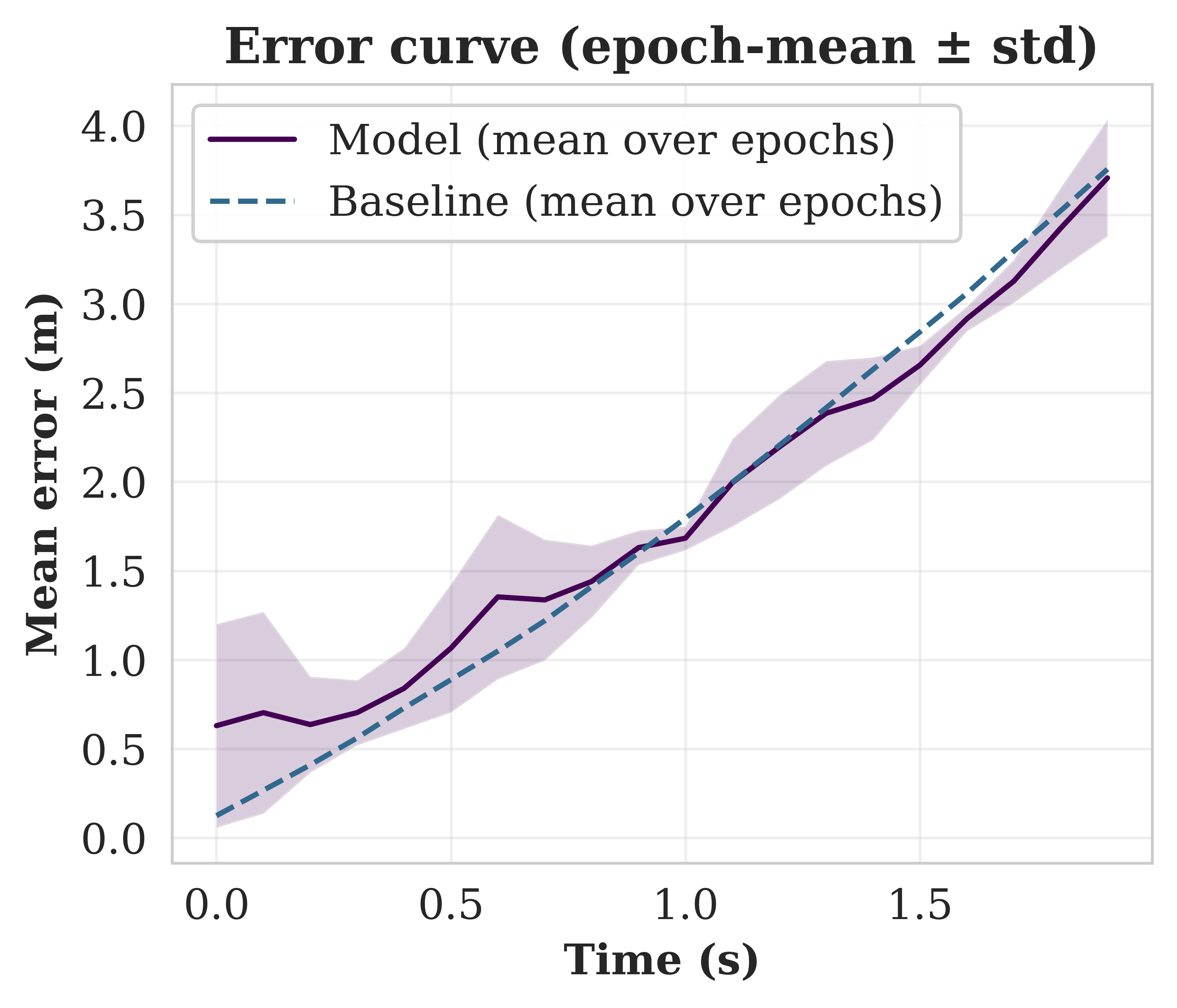}
    \caption{Error over horizon (epoch-mean $\pm$ std).}
    \label{fig:acc-error-curve}
  \end{subfigure}

  \caption{Accuracy overview across training. \textbf{(a)} ADE/FDE drop quickly in the first 10–15 epochs then taper, indicating fast capture of short-horizon kinematics with later refinement. \textbf{(b)} Miss@\SI{2}{m} and Miss@\SI{4}{m} decrease and stabilize. \textbf{(c)} Percentile ADE (P50–P99) shows substantial shrinkage of the bulk and mid-tail, with little change in the extreme tail. \textbf{(d)} P90–P100 and P95–P100 bands narrow steadily, indicating improved error concentration. \textbf{(e)} ADE vs.\ lane/CTRV baseline: the model opens and sustains a gap over the kinematic prior. \textbf{(f)} Horizon-wise error curves over \SI{0}{s}–\SI{2.0}{s} show a consistent advantage across time.}

  \label{fig:accuracy-grid}
  \label{fig:accuracy-time}
  \label{fig:accuracy-percentiles}
  \label{fig:acc-kinematic-pair}
\end{figure*}

\paragraph{Global loss behaviour.}
Training and validation losses decrease smoothly over 100 epochs as shown in Fig.~\ref{fig:global-loss}, with the validation curve closely tracking the training curve, evidence that the shallow, 9-qubit circuits generalize without overfitting. Two design choices make this possible: (i) predicting \emph{residuals} in a lane-aligned frame compresses output range and removes most rigid-body motion; (ii) feature normalization bounds all rotation angles, keeping SPSA perturbations well-conditioned.

\paragraph{Fine-grained learning signals.}
The per-step traces reveal how the hybrid system progresses. The raw step-level loss and its smoothed trajectory (Fig.~\ref{fig:signals-loss-dynamics}, Fig.~\ref{fig:signals-step-loss}) show steady improvement with gentle undulations that coincide with the epoch-wise SPSA schedule resets. The finite-difference loss rate (Fig.~\ref{fig:signals-loss-rate}) oscillates around a negative mean, which is characteristic of objectives with a $\min$ over modes: when a new hypothesis becomes the best explainer for a scene, the local slope changes sign but the global trend remains downward. Step-level ADE/FDE (Fig.~\ref{fig:signals-step-errors}) steadily contract in amplitude, indicating that volatility becomes localized to a shrinking subset of challenging scenes.
\paragraph{Convergence and stability.}
A 'best-so-far' validation ADE forms a staircase that drops at epochs $\sim$6, 18, 35, 52, 74 and 83 (\ref{fig:conv-035}); those steps align with parameter regimes where the shallow attention and decoder phases re-align to the Fourier readout. The rolling standard deviation of ADE collapses by an order of magnitude from early to late training (\ref{fig:conv-034}), and the epoch-to-epoch improvement trend tapers to near-zero after $\sim$70 epochs (\ref{fig:conv-032}), both consistent with well-behaved convergence. In parallel, variance over training settles to a tight band (\ref{fig:conv-031}). Together these diagnostics show that the shallow circuits admit stable optimization and predictable convergence under SPSA.

\subsection{Accuracy over time and against a kinematic baseline}

\paragraph{ADE/FDE across epochs and miss rates.}
Average and terminal errors decline rapidly in the first 10–15 epochs and then plateau gracefully (Fig.~\ref{fig:acc-ade-fde}). Miss rates track this improvement: Miss@\SI{2}{m} and Miss@\SI{4}{m} drop steadily and stabilize (Fig.~\ref{fig:acc-miss}). The pattern is mechanistically consistent with our design: the attention encoder quickly captures short-horizon kinematics, while the Fourier decoder refines terminal placement via phase adjustments.

\paragraph{Percentile structure of errors.}
Error percentiles reveal \emph{where} the gains accrue. The high-volume region improves markedly, P90 ADE falls to approximately \SIrange{3.5}{4.0}{m} and P95 to approximately \SIrange{9}{10}{m}, whereas P99 remains approximately constant (Fig.~\ref{fig:acc-percentiles}). An alternate percentile progression view (Fig.~\ref{fig:acc-percentile-bands}) confirms the same trend with shaded P90-P100 and P95-P100 bands. This profile is exactly what a truncated Fourier basis ($B{=}8$) should produce: strong bias toward smooth, lane-conforming adjustments (head and mid-tail), with a small long-tail associated with abrupt late decisions.

\paragraph{Comparison to a strong kinematic baseline.}
Against a lane/CTRV extrapolation, the model’s ADE curve closes the gap epoch by epoch (Fig.~\ref{fig:acc-kinematic}), and a time-profiled error curve averaged across epochs shows a consistent advantage across the prediction horizon (Fig.~\ref{fig:acc-error-curve}). Importantly, our approach delivers this accuracy \emph{and} a multi-hypothesis distribution in a single quantum pass-an efficiency benefit inherent to the superposition-based decoder.

\begin{figure*}[t]
  \centering
  \begin{subfigure}[t]{0.3\textwidth}
    \centering
    \includegraphics[width=0.9\linewidth]{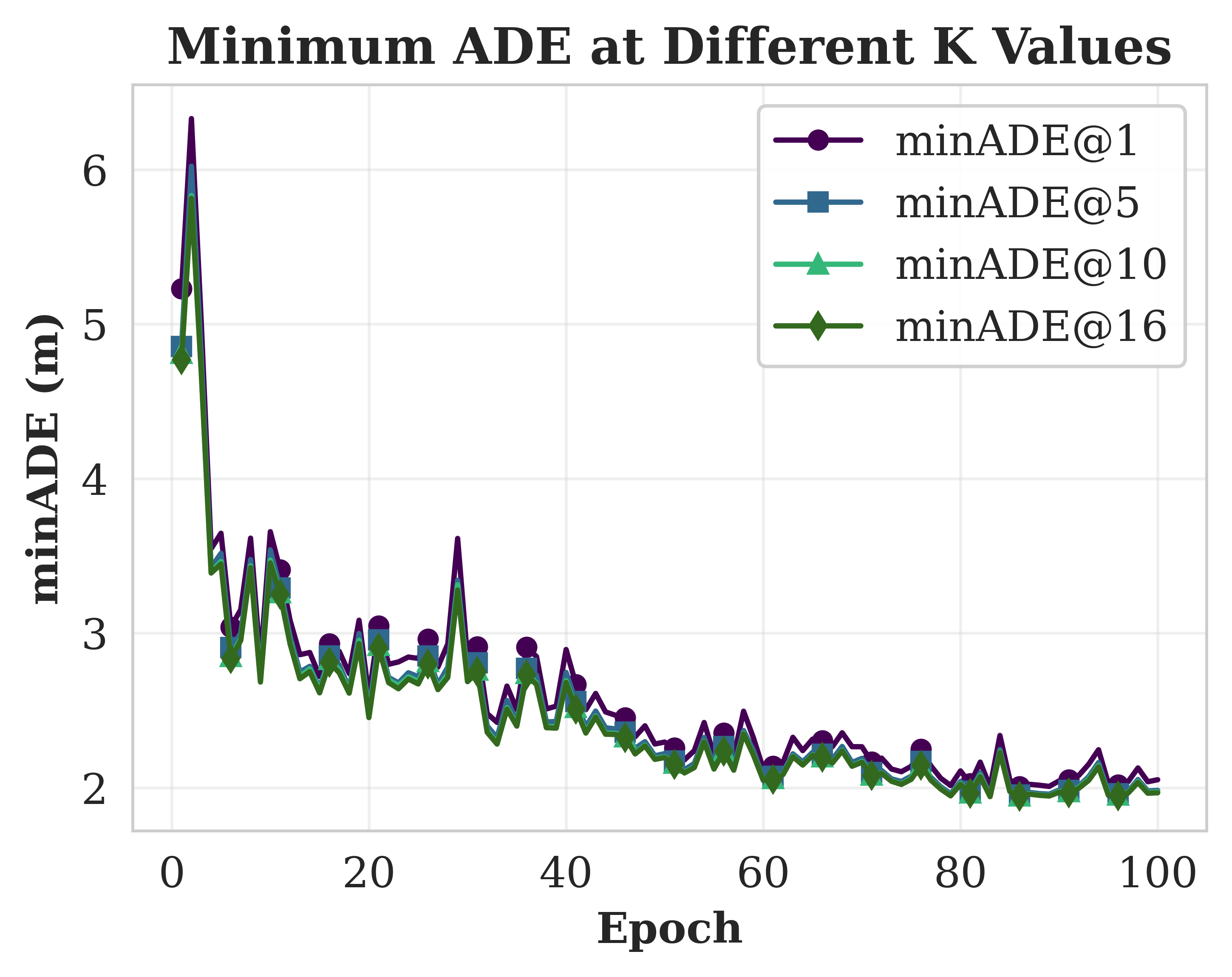}
    \caption{Min ADE vs.\ top-$K$ (confidence-ranked).}
    \label{fig:mm-min-ade}
  \end{subfigure}\hfill
  \begin{subfigure}[t]{0.3\textwidth}
    \centering
    \includegraphics[width=0.9\linewidth]{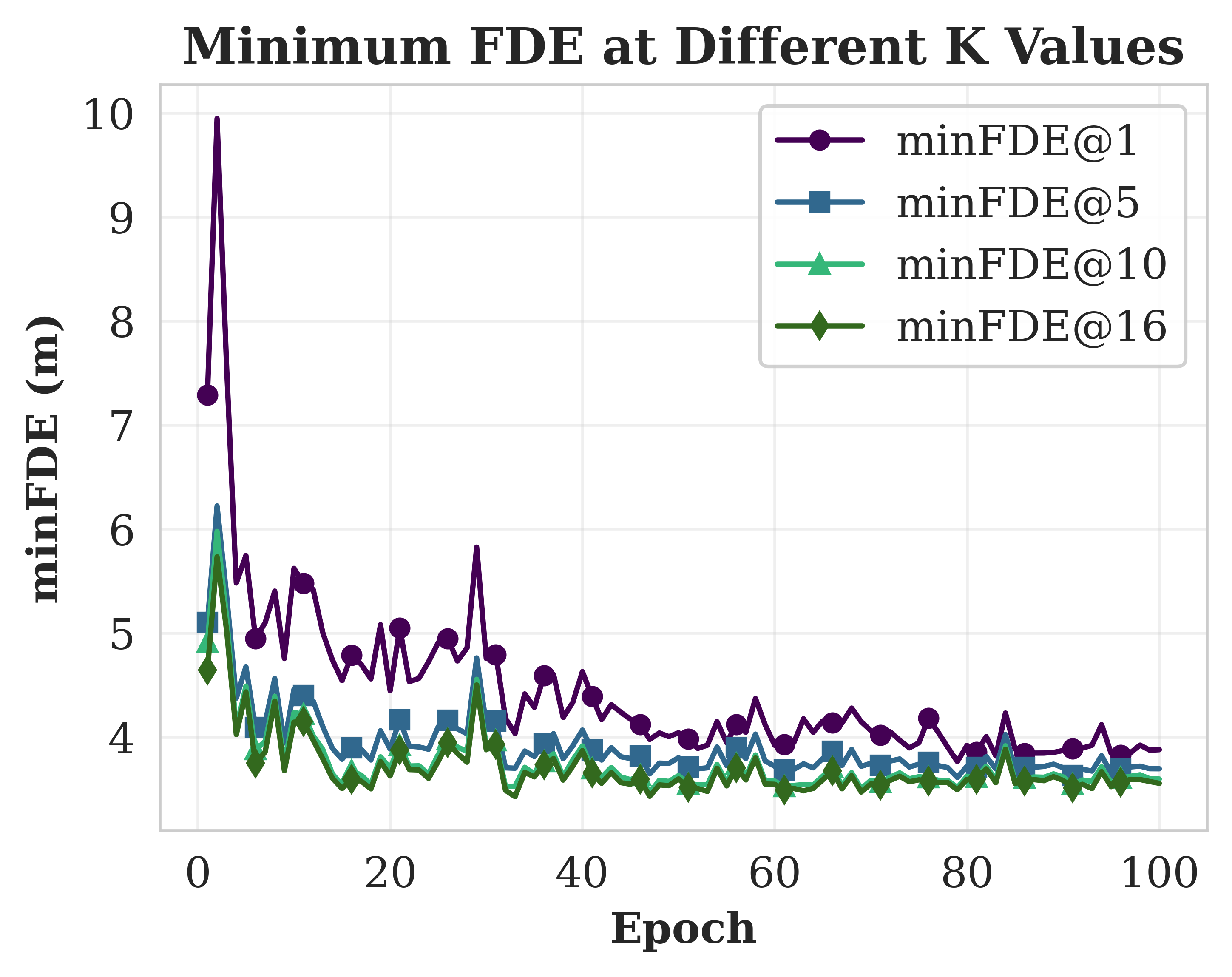}
    \caption{Min FDE vs.\ top-$K$ (confidence-ranked).}
    \label{fig:mm-min-fde}
  \end{subfigure}\hfill
  \begin{subfigure}[t]{0.3\textwidth}
    \centering
    \includegraphics[width=0.9\linewidth]{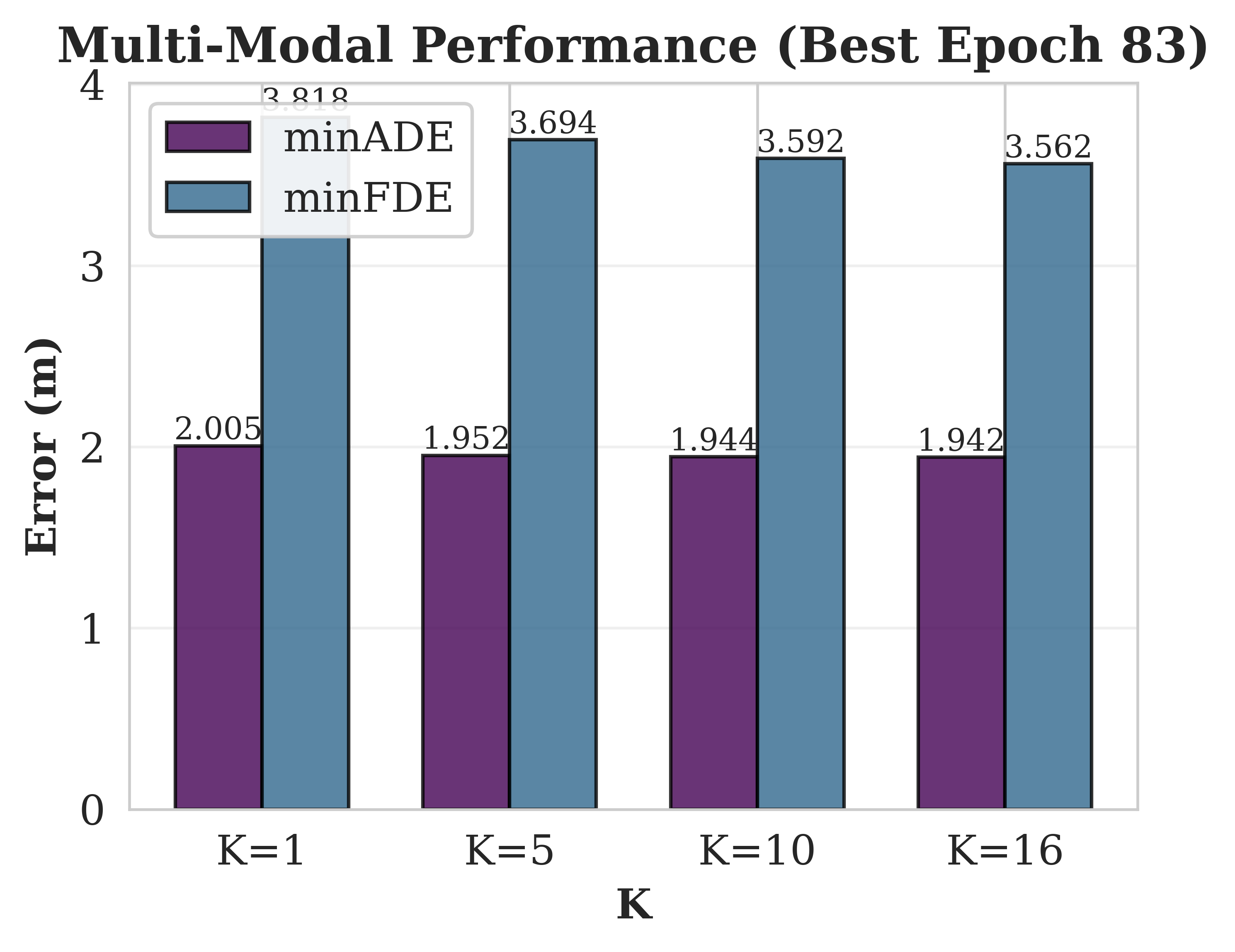}
    \caption{Selection epoch (83): key values for $K{=}5$ and $K{=}16$.}
    \label{fig:mm-best-epoch-metrics}
  \end{subfigure}

  \caption{Minimum ADE/FDE over the top-$K$ confidence-ranked hypotheses decrease monotonically with $K$, with shallow slopes—indicating that even lower-ranked modes remain close to the best hypothesis. At the model-selection epoch (83), values reported in the text are read from the right panel.}
  \label{fig:mm-min-errors}
\end{figure*}

\begin{figure*}[t]
  \centering
  \begin{subfigure}[t]{0.3\textwidth}
    \centering
    \includegraphics[width=0.9\linewidth]{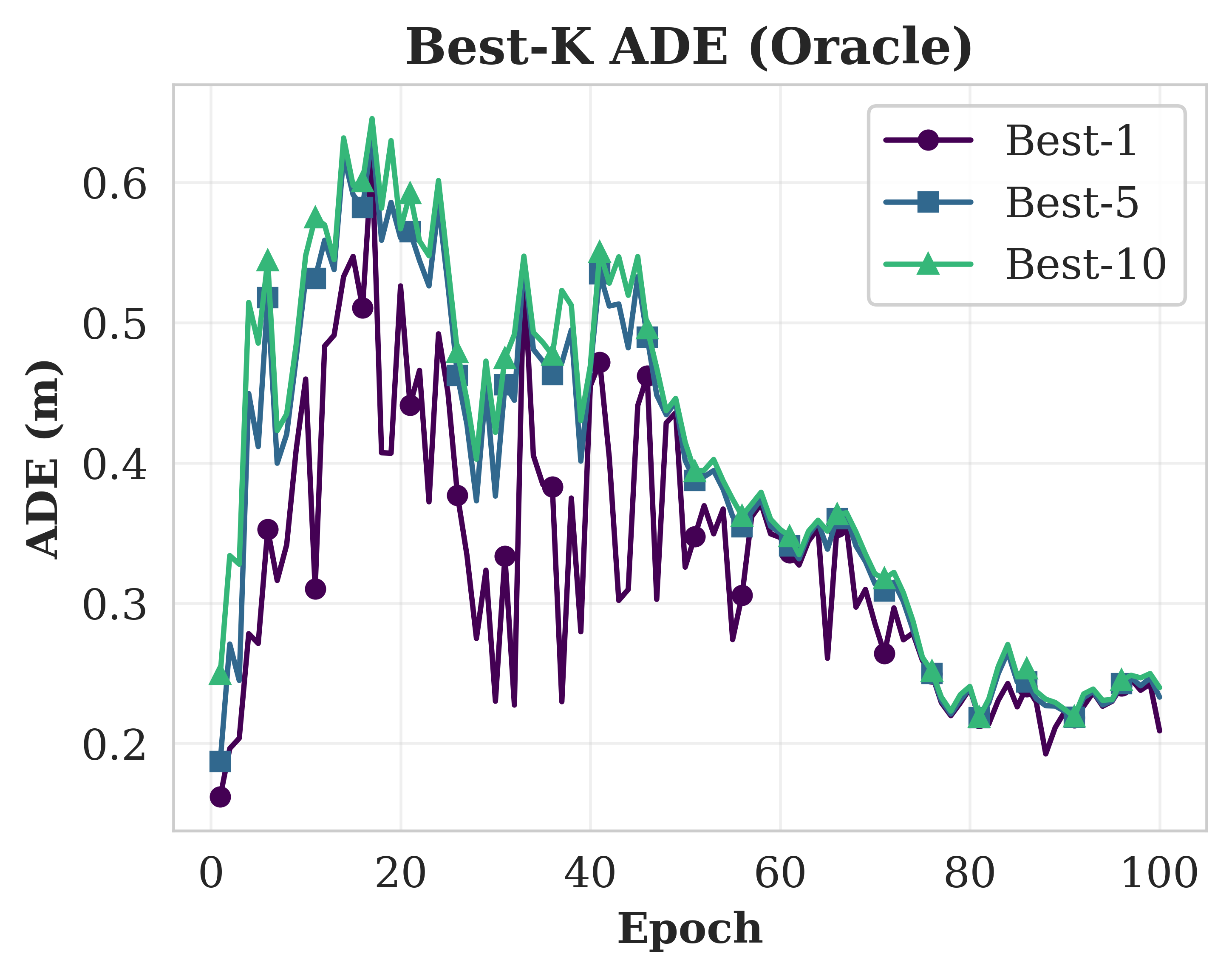}
    \caption{Best-$K$ (oracle) ADE over epochs.}
    \label{fig:mm-oracle-ade}
  \end{subfigure}\hfill
  \begin{subfigure}[t]{0.3\textwidth}
    \centering
    \includegraphics[width=0.9\linewidth]{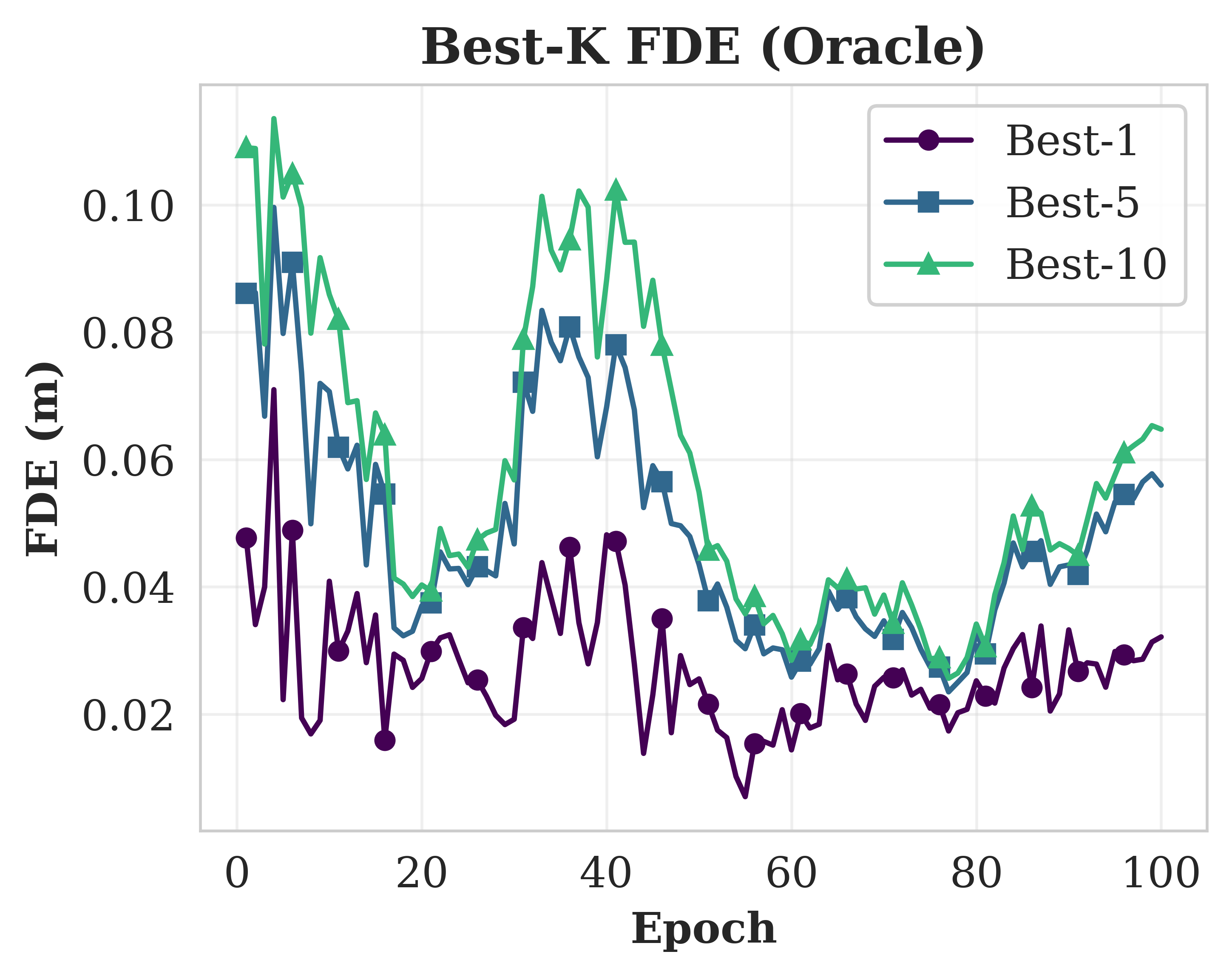}
    \caption{Best-$K$ (oracle) FDE over epochs.}
    \label{fig:mm-oracle-fde}
  \end{subfigure}\hfill
  \begin{subfigure}[t]{0.3\textwidth}
    \centering
    \includegraphics[width=0.9\linewidth]{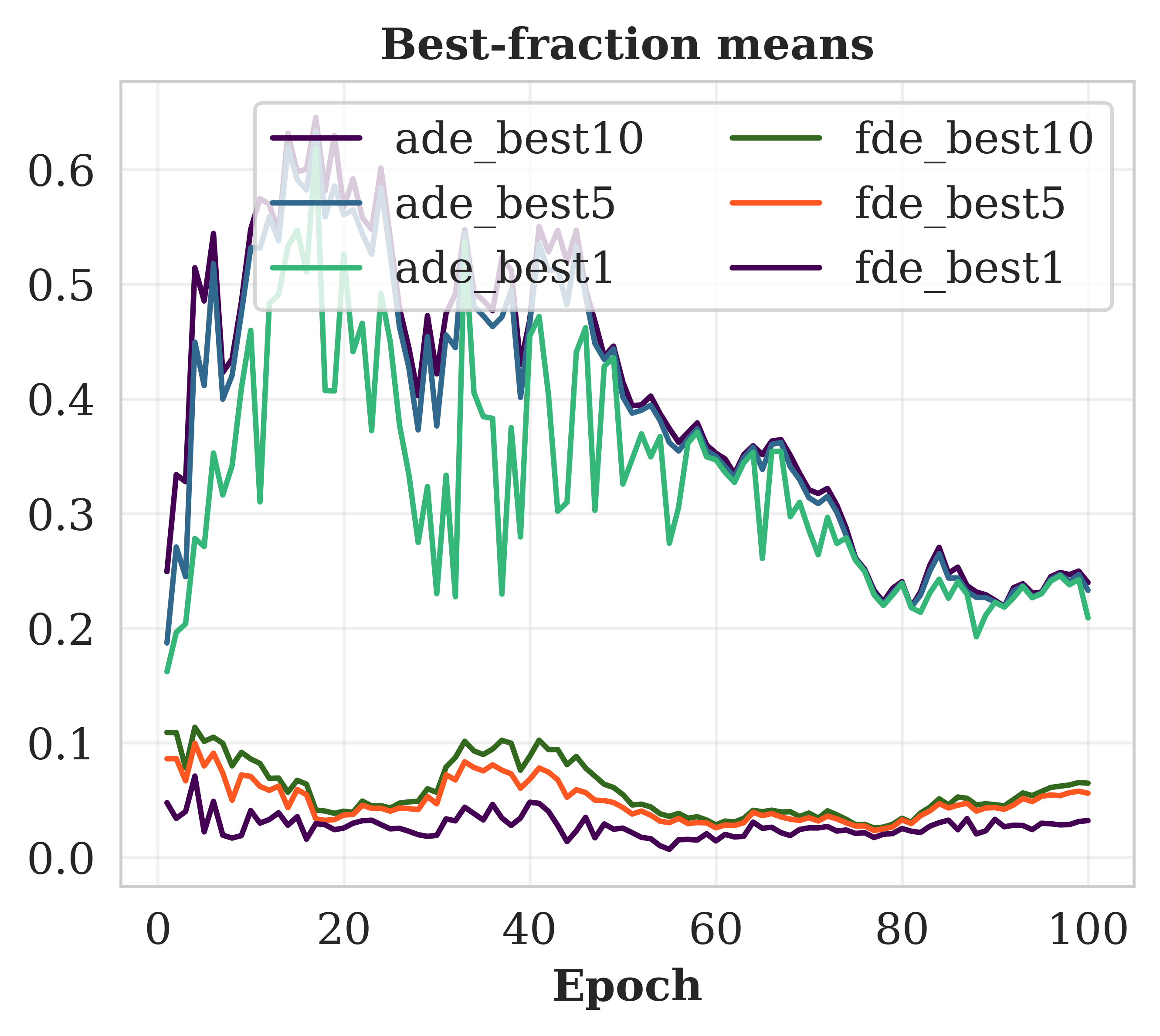}
    \caption{Best-$K$ means (ADE upper cluster, FDE lower cluster).}
    \label{fig:mm-best-fraction}
  \end{subfigure}

  \caption{Oracle set quality across training. Left/middle: Best-$K$ ADE and FDE over epochs. Right: compact summary of ADE/FDE Best-$K$ means for $K\in\{1,5,10\}$ showing a brief warm-up (epochs $\sim$5–20), steady decline, and taper after $\sim$70 epochs (meters after de-normalization).}
  \label{fig:mm-oracle}
\end{figure*}

\begin{figure*}[t]
  \centering

  \begin{subfigure}[t]{0.24\textwidth}
    \centering
    \includegraphics[width=\linewidth]{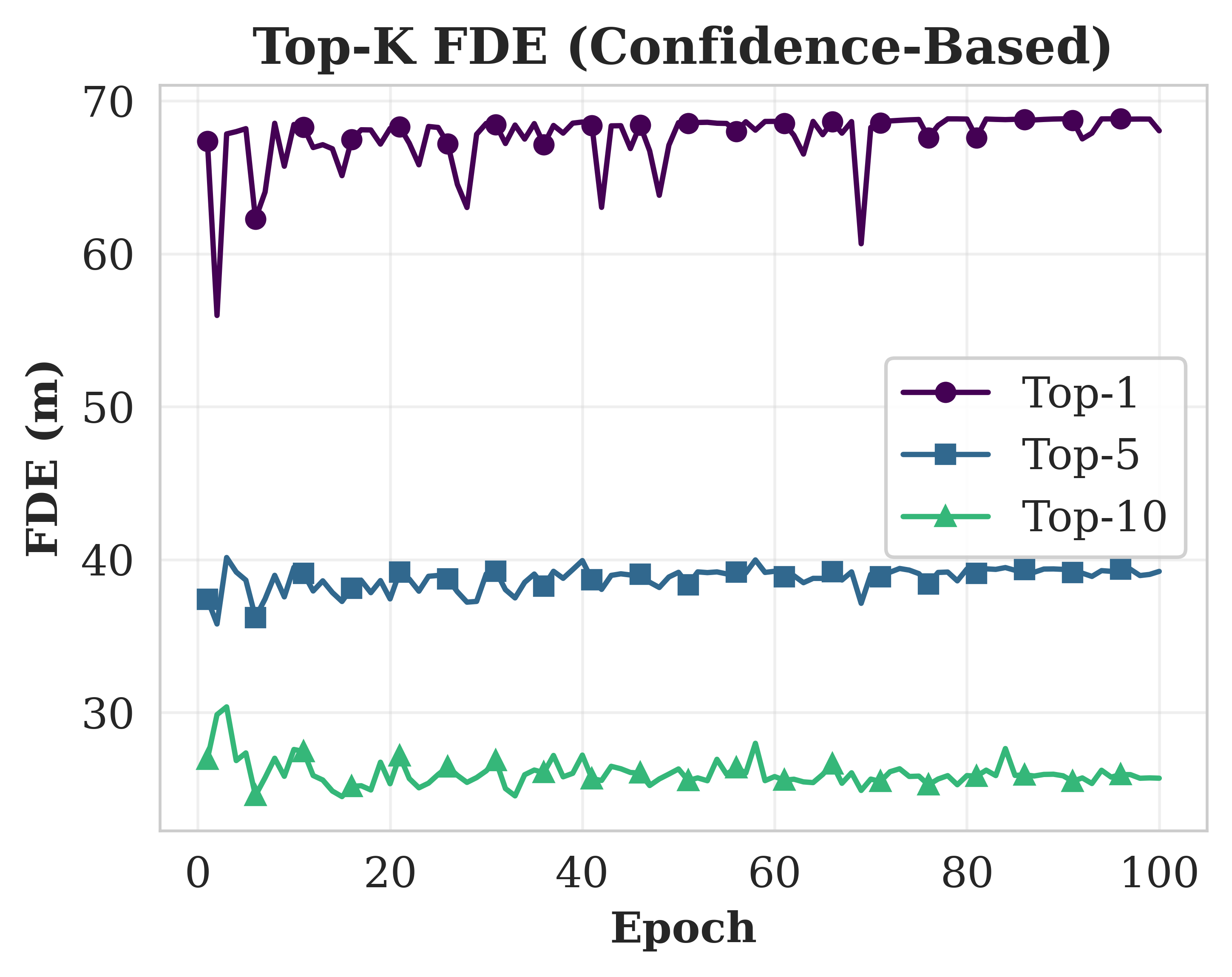}
    \caption{Top-$K$ FDE across training.}
    \label{fig:mm-topk-fde}
  \end{subfigure}\hfill
  \begin{subfigure}[t]{0.24\textwidth}
    \centering
    \includegraphics[width=\linewidth]{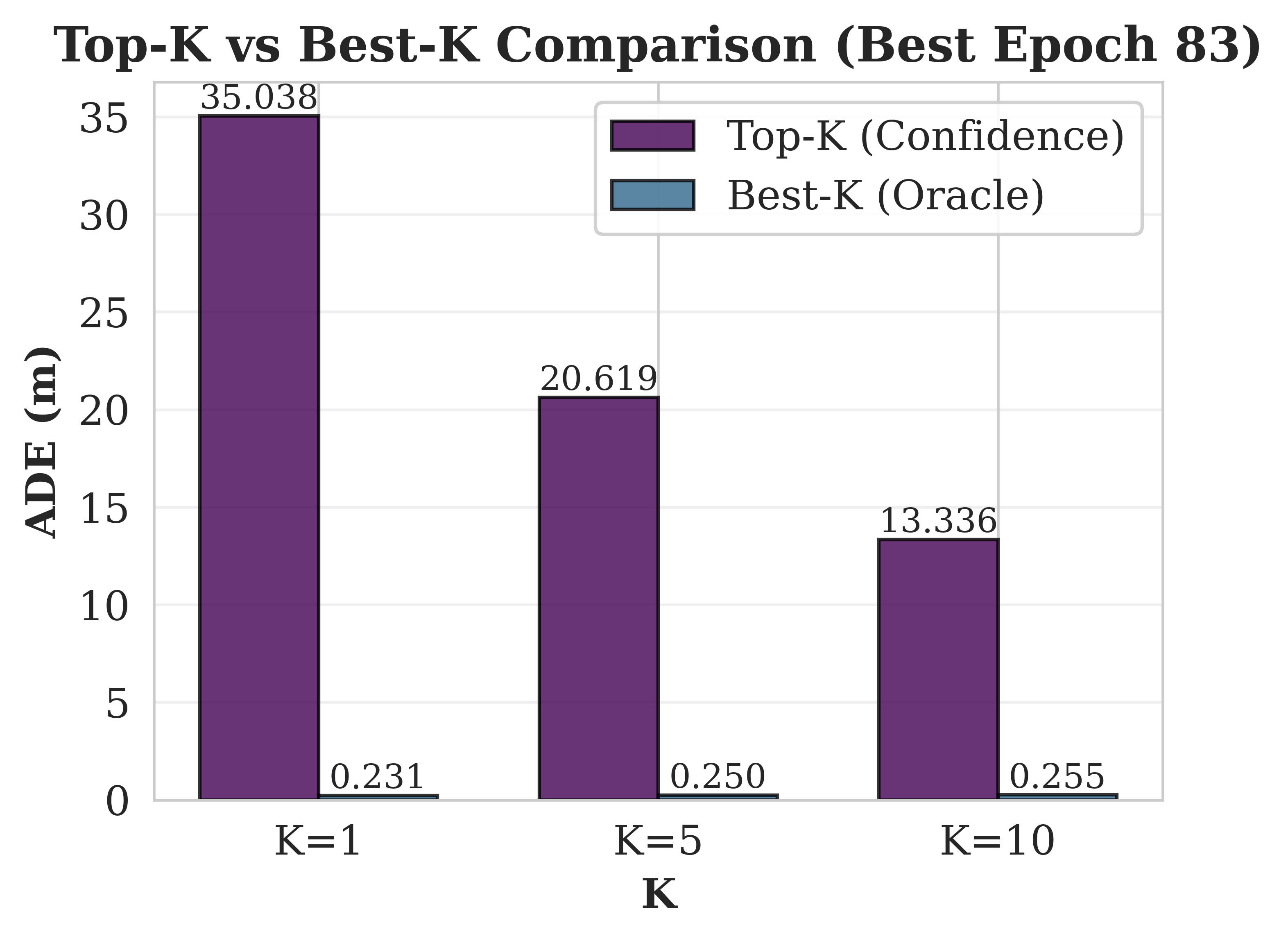}
    \caption{Epoch 83: Top-$K$ vs.\ Best-$K$ ADE.}
    \label{fig:mm-topk-vs-bestk}
  \end{subfigure}\hfill
  \begin{subfigure}[t]{0.24\textwidth}
    \centering
    \includegraphics[width=\linewidth]{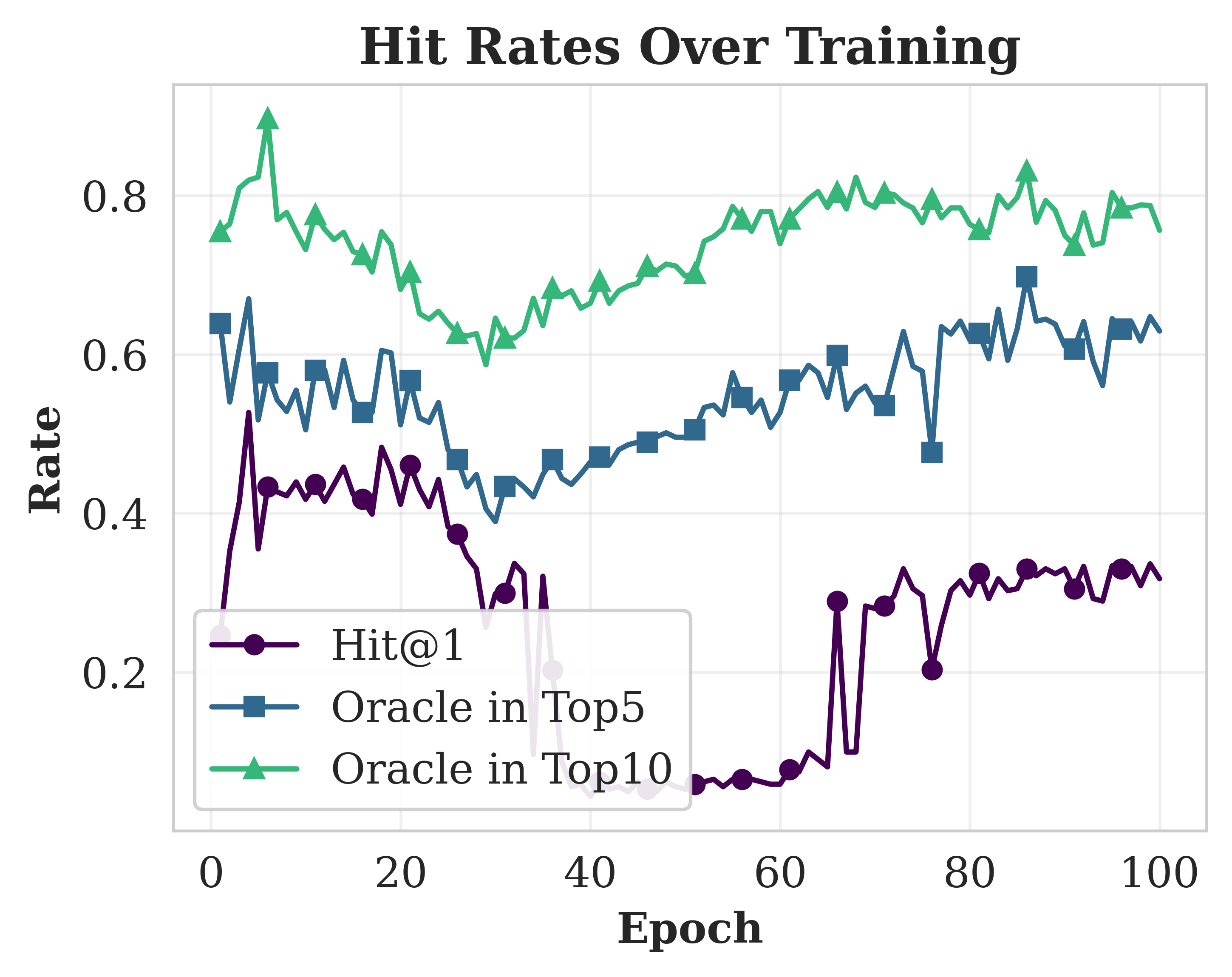}
    \caption{Hit@1 and oracle-in-Top-$K$.}
    \label{fig:hit-rates}
  \end{subfigure}\hfill
  \begin{subfigure}[t]{0.24\textwidth}
    \centering
    \includegraphics[width=\linewidth]{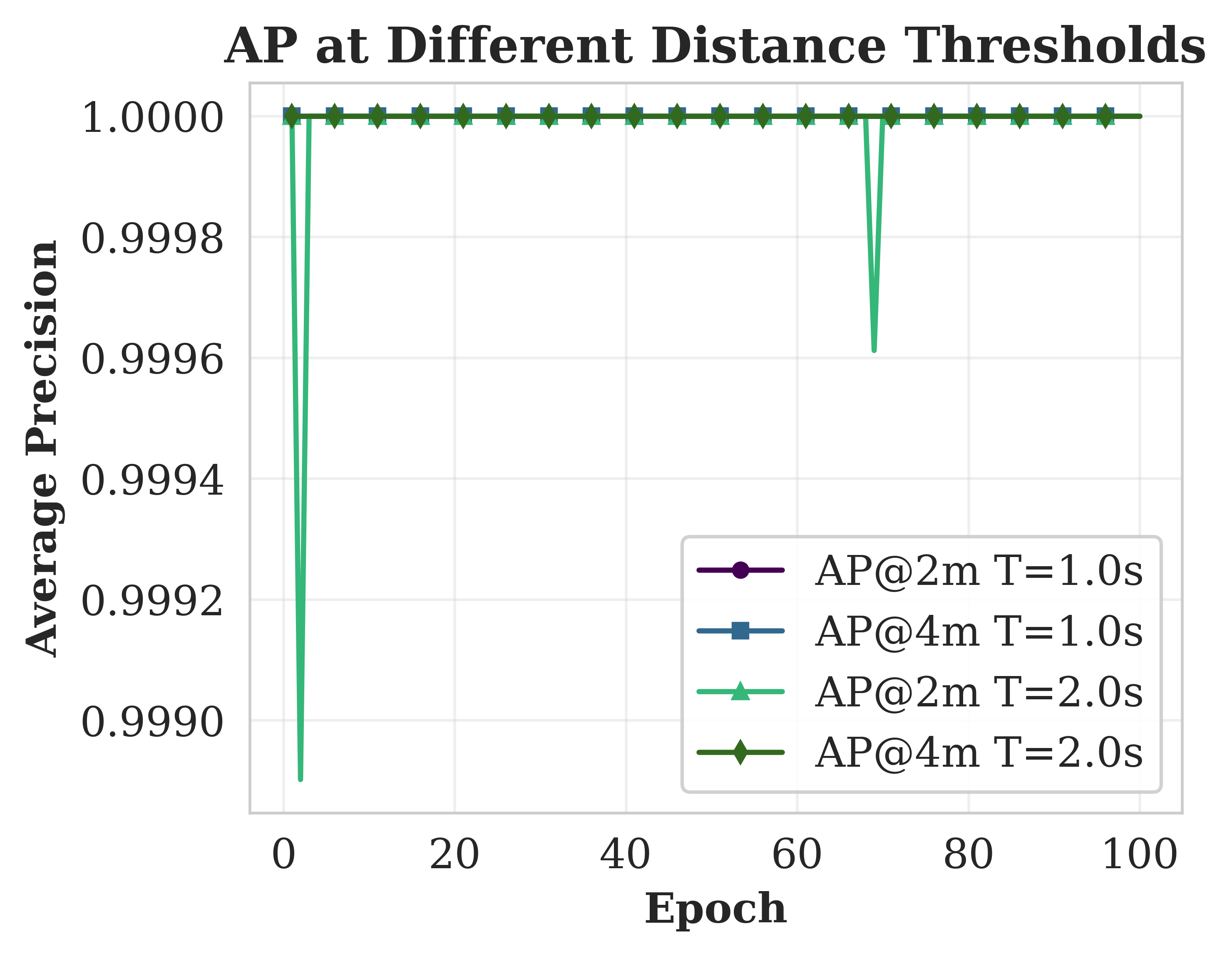}
    \caption{Average Precision across epochs.}
    \label{fig:ap-thresholds-epochs}
  \end{subfigure}
\caption{Ranking, dispersion, and calibration overview. \textbf{(a)} Top-$K$ FDE (confidence-ranked) decreases with $K$ and remains stable across training, indicating controlled long-horizon dispersion. \textbf{(b)} At the selection epoch (83), the gap between Top-$K$ (ranked) and Best-$K$ (oracle) ADE measures representational headroom, with ranking already capturing much of it at modest $K$. \textbf{(c)} Hit@1 rises over training while 'oracle in Top-$K$' rates stay high, indicating strong ranking quality. \textbf{(d)} AP stays consistently high across epochs, summarizing precision–recall behaviour under the adopted evaluation protocol.}
  \label{fig:ranking-dispersion-grid}
\end{figure*}

\subsection{Multi-modal generation: coverage and efficiency}

\paragraph{Minimum errors over the top-$K$ ranked hypotheses.}
The minimum ADE and FDE measured over the top-$K$ confidence-ranked modes decrease monotonically with $K$ (Fig.~\ref{fig:mm-min-ade}, Fig.~\ref{fig:mm-min-fde}); at the model-selection epoch (83), the values are \SI{1.952}{m}/\SI{3.694}{m} for $K{=}5$ and \SI{1.942}{m}/\SI{3.562}{m} for $K{=}16$ (Fig.~\ref{fig:mm-best-epoch-metrics}). The shallow slope indicates that even lower-ranked modes remain close to the best hypothesis, consistent with the decoder’s coherent phase-family construction.

\paragraph{Oracle set quality.}
Best-$K$ (oracle) curves evidence the quality of the hypothesis \emph{set}. Best-$K$ ADE and FDE fall well below the corresponding Top-$K$ values in later epochs (Fig.~\ref{fig:mm-oracle-ade}, Fig.~\ref{fig:mm-oracle-fde}; see also Fig.~\ref{fig:mm-best-fraction}, which compacts the ADE/FDE Best-$K$ means across $K\in\{1,5,10\}$ and shows a brief warm-up (epochs $\sim$5–20) followed by a steady decline that tapers after $\sim$70 epochs). This is a key result: with only 9 qubits and shallow entanglement, a single forward pass routinely \emph{contains} at least one trajectory that lies close to the realized future among its superposed modes.

\paragraph{Top-$K$ end-point dispersion.}
Top-$K$ FDE measured on the highest-confidence hypotheses exhibits stable behaviour across training and decreases with $K$ (Fig.~\ref{fig:mm-topk-fde}). The level and smoothness of these curves reflect consistent long-horizon end-point dispersion under a stringent measure applied uniformly across heterogeneous scenes.

\paragraph{Top-$K$ vs.\ Best-$K$ at the selection epoch.}
A compact bar plot contrasts Top-$K$ (confidence-ranked) and Best-$K$ (oracle) ADE at epoch 83 (Fig.~\ref{fig:mm-topk-vs-bestk}). The oracle bars quantify the \emph{representational headroom} available from the same shallow circuits; the Top-$K$ bars confirm that the chosen ranking still captures a large share of that headroom at modest $K$. This gap is useful: it indicates that further improvements can be achieved by purely algorithmic changes to scoring, without modifying circuit depth or qubit count.

\begin{figure}[t]
  \centering
  \includegraphics[width=0.85\columnwidth]{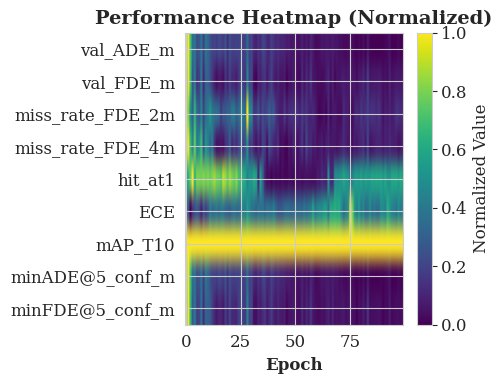}
  \caption{Temporal overview: normalized performance heatmap over 100 epochs showing coherent evolution across metrics.}
  \label{fig:performance-heatmap}
\end{figure}

\subsection{Ranking quality, recall and calibration}

\paragraph{Hit rates and recall.}
Hit@1 increases over training while the “oracle within top-$K$” rates remain high (Fig.~\ref{fig:hit-rates}). Across horizons (\SI{1.0}{s}, \SI{2.0}{s}), thresholds (\SI{2}{m}, \SI{4}{m}), and $K\in\{1,5,10,16\}$, recall stays consistently strong, and at the selection epoch coverage is near-uniform across these settings.

\paragraph{Calibration and precision–recall summaries.}
Expected Calibration Error evolves gradually and remains stable across epochs, as seen in the ECE row of the temporal performance heatmap (Fig.~\ref{fig:performance-heatmap}), consistent with our deterministic, scene-level confidence mapping derived from the latent spectrum. Scene-wise mAP remains nearly constant, and Average Precision stays high across epochs (Fig.~\ref{fig:ap-thresholds-epochs}); at the selection epoch, AP at the standard distance thresholds is effectively saturated.

\subsection{Aggregated views and metric relations}

\paragraph{All metrics at the selection epoch.}
At the selection epoch, the aggregate snapshot shows meter-scale ADE, multi-meter FDE, and further reductions for \textit{minADE}/\textit{minFDE} at modest $K$. Miss@\SI{2}{m} and Miss@\SI{4}{m} are substantially lower than for the kinematic baseline, Hit@1 is improved, and calibration/precision–recall measures remain stable, indicating strong accuracy and broad hypothesis coverage from a lightweight quantum model. Although we do not aim for state-of-the-art performance, it is useful to contextualize the numbers: recent large classical models on the WOMD typically report ADE/FDE in the sub-meter to \(\sim\)1–2\,m range for comparable short-horizon settings, using orders of magnitude more parameters and richer scene encoders. Our 9-qubit, \(\sim\!10^3\)-parameter model sits above these best-in-class error levels, but does so while improving over a strong kinematic baseline and providing single-pass multi-modality in a hardware-aligned quantum formulation.

\paragraph{Temporal overview.}
A normalized performance heatmap aligns the trajectories of all major metrics across the 100 epochs (Fig.~\ref{fig:performance-heatmap}): the ADE/FDE rows progressively darken (improve), miss rates cool in tandem, and Hit@1 brightens as training proceeds, showing coherent evolution across metrics from a single training run.


\subsection{Why shallow circuits are effective here}
The results above arise from pairing \emph{task structure} with \emph{quantum inductive bias}. Residual learning in an ego-centric lane frame ensures the quantum layers model only small, structured deviations, well within the expressivity of 9 qubits with linear/ladder entanglement. The Fourier decoder encodes a smoothness prior matched to short-horizon driving statistics: it captures the bulk of trajectories at very low circuit depth and, via global phase superposition, produces multiple plausible futures without additional passes. SPSA with bounded rotations then yields stable optimization despite the non-smooth “min over modes” objective, as reflected in the loss, variance and convergence diagnostics (Fig.~\ref{fig:signals-loss-dynamics}, Fig.~\ref{fig:signals-step-loss}, Fig.~\ref{fig:signals-loss-rate}, Fig.~\ref{fig:conv-034}, Fig.~\ref{fig:conv-035}).

\paragraph{Resource-efficient design that drives accuracy.}
Our performance gains stem from design choices that shrink the function class the quantum circuit must represent while preserving the behaviours that matter for short-horizon driving. Predicting \emph{residuals} in an ego-lane frame atop a map/CTRV baseline means the circuit models only small, smooth corrections instead of full trajectories, concentrating probability mass near lane-consistent futures and explaining the rapid ADE/FDE decline and lower slope of the error-over-time curves relative to the baseline (Fig.~\ref{fig:acc-kinematic}). The decoder’s truncated Fourier basis with phase superposition simultaneously imposes a low-frequency smoothness prior and emits multiple coherent hypotheses in a \emph{single} pass, yielding strong Recall@K and stable mAP without scaling inference cost with $K$ (summarized above; see Fig.~\ref{fig:mm-best-epoch-metrics}). Shallow entanglement with per-qubit rotations keeps the parameter count tiny and the SPSA landscape well-conditioned, producing the clean convergence and shrinking variance we observe (Fig.~\ref{fig:signals-loss-dynamics}, Fig.~\ref{fig:conv-034}, Fig.~\ref{fig:conv-031}). Finally, confidences derived from the latent spectrum (FFT of the decoded state) mean ranking quality emerges from the same physics-shaped representation that generates trajectories, explaining the strong Top-$K$ curves and high 'oracle in Top-5/10' rates without an extra learned head (Fig.~\ref{fig:mm-best-epoch-metrics}, Fig.~\ref{fig:hit-rates}). Collectively, these design decisions allow a 9-qubit, shallow-depth model to deliver meter-level accuracy, broad hypothesis coverage and stable training, achieving useful performance while remaining computationally and parameter efficient.

\section{Discussion}
\label{sec:discussion}

\subsection{Complexity, expressivity and multi-modality}
\label{sec:complexity-expressivity}

The decoder operates on $n{=}9$ qubits, i.e.\ a $2^n{=}512$-dimensional Hilbert space, while the full model uses only $N_{\theta}\!\approx\!1.2{\times}10^3$ trainable angles across attention, feedforward and decoder circuits. From a classical viewpoint, compositions of trigonometric rotations and sparse entanglement implement a rich non-linear feature map over the SDV history, with the truncated Fourier readout acting as a compact linear head in a low-frequency basis. Oracle metrics expose the effective capacity of this setup: Best-$K$ (oracle) ADE/FDE curves lie well below their Top-$K$ (confidence-ranked) counterparts for moderate $K$ (e.g.\ $K\in\{5,10,16\}$), showing that in many scenes at least one of the $M{=}16$ hypotheses lies close to the realized future. The remaining gap is largely due to ranking rather than representation—the shallow circuits can already place some modes near ground truth within the truncated Fourier basis, but the spectrum-derived confidences do not always surface these modes at the very top. The multi-modal construction is deliberately frugal: all $M$ hypotheses share a single decoder and differ only by global phase offsets applied before Fourier readout. This 'phase sweep' produces a correlated family of futures that share a smooth backbone but diverge along a few phase directions, without increasing depth or qubit count, yielding a hypothesis set that is expressive for short-horizon motion while keeping ranking simple and hardware-aligned.

\subsection{Relationship to classical baselines and large models}

Relative to the lane/CTRV baseline, the model opens and maintains an ADE/FDE gap across the prediction horizon, while also reducing miss rates and improving Hit@1 and related recall measures. This indicates that the quantum components learn useful residual structure beyond the deterministic prior, even under tight parameter and depth budgets, with the largest gains appearing in the bulk and mid-tail of the scene distribution where the lane-aligned residual and Fourier priors are most appropriate. Absolute error levels remain above those of large classical GNN and transformer architectures on WOMD, which is expected given that such models employ orders of magnitude more parameters, ingest full multi-agent and HD-map context, and are trained end-to-end with backpropagation. The goal here is not to match those systems, but to show that a small quantum sub-module, designed for NISQ constraints, can train stably, produce coherent multi-modal outputs and reliably outperform a strong kinematic baseline. From a deployment perspective, the single-pass multi-modality of the Fourier decoder is particularly attractive: all $M$ trajectories are generated in one quantum forward pass and confidences are derived from the same latent spectrum, so decoding cost is effectively constant in $M$, in contrast to classical multi-head or iterative schemes whose inference time scales with the number of modes.

\subsection{Limitations and scope}

This study has important limitations. All experiments are run in classical simulation, so noise, decoherence and device-specific constraints are not captured; although the circuits are shallow and low-parameter to be NISQ-friendly, the reported metrics do not yet reflect real hardware behaviour. The formulation is SDV-centric with strong preprocessing: the lane-aligned residual representation compresses the prediction problem but omits rich interaction patterns between agents and much of the HD-map structure beyond local lane direction. Consequently, highly interactive or rare manoeuvres such as merges, near-collisions and abrupt lane changes remain challenging for a low-frequency Fourier basis, which is visible in the persistent long tail of large ADE/FDE cases. The evaluation further targets a short (2\,s) horizon on a large but still limited subset of WOMD due to simulation cost, and the work does not claim quantum advantage over strong classical forecasters. Instead, it should be viewed as a controlled feasibility study: under realistic resource constraints, shallow, few-qubit circuits can be integrated into a physics-shaped pipeline and trained to deliver non-trivial, multi-modal trajectory forecasts on a modern autonomous-driving benchmark.

\section{Conclusion and Future Work}
\label{sec:conclusion}

This paper presented a compact, physics-shaped hybrid quantum architecture for short-horizon trajectory forecasting in autonomous driving. By working in an ego-centric, lane-aligned frame and predicting residuals on top of a map/CTRV baseline, the model focuses on smooth, meter-scale corrections instead of full trajectories. A transformer-inspired quantum attention encoder, a deep but parameter-lean feedforward stack, and a Fourier-based decoder together form a 9-qubit, $\sim\!10^3$-parameter pipeline that generates $M{=}16$ trajectory hypotheses in a single pass, with confidences derived deterministically from the latent spectrum.

On a representative subset of the WOMD, the model attains meter-scale ADE/FDE, improves consistently over a strong kinematic baseline, and shows stable miss rates, Hit@1, and precision–recall behaviour across training. Multi-modal metrics indicate that, for many scenes, at least one of the 16 hypotheses lies close to the realized future, and that the simple spectrum-based ranking captures a substantial fraction of this representational headroom. Training with SPSA remains stable despite non-differentiable components, demonstrating that shallow, few-qubit circuits can be optimized reliably in this setting.

The study remains limited to classical simulation, a single-SDV formulation with strong preprocessing, and a short prediction horizon. Future work includes deploying the architecture on near-term hardware, incorporating richer multi-agent and HD-map context, exploring more expressive or adaptive quantum decoders, and refining ranking and calibration mechanisms. Beyond autonomous driving, the residual, physics-shaped formulation and phase-based multi-modality explored here are applicable to other resource-constrained sequential prediction problems in robotics, logistics, and human motion forecasting.

\bibliographystyle{ieeetr}
\bibliography{main}

\end{document}